\documentclass[lettersize,journal]{IEEEtran}
\usepackage{amsmath,amsfonts}
\usepackage{algorithmic}
\usepackage{algorithm}
\usepackage{array}
\usepackage[caption=false,font=normalsize,labelfont=sf,textfont=sf]{subfig}
\usepackage{textcomp}
\usepackage{stfloats}
\usepackage{url}
\usepackage{verbatim}
\usepackage{graphicx}
\usepackage{cite}
\usepackage{makecell}
\usepackage{multirow}
\usepackage{xspace}
\usepackage{amsmath}
\usepackage{amssymb}
\usepackage{mathtools}
\usepackage{amsthm}
\usepackage{booktabs} 
\usepackage[capitalize,noabbrev]{cleveref}
\usepackage{color}
\usepackage{xcolor} 
\usepackage{colortbl}
\theoremstyle{plain}

\theoremstyle{definition}

\theoremstyle{remark}

\def \E {\mathrm{E}}
\def \x {\mathbf{x}}

\def \L {\mathcal{L}}
\def \D {\mathcal{D}}

\def \R {\mathbb{R}}

\def \N {\mathcal{N}}

\def \P {\mathbb{P}}

\def \P {\mathcal{P}}
\def \diag {\mbox{diag}}
\definecolor{Gray}{gray}{0.85}
\newcommand{\Gray}[0]{\rowcolor{gray!20}}
\definecolor{sclgreyblue}{rgb}{0.2,0.3,0.5}%

\newcommand{\abbr}[0]{OneNet\xspace}
\newcommand{\abbrv}[0]{D$^3$A\xspace}
\newtheorem{thm}{Theorem}
\newtheorem{prop}{Proposition}

\usepackage[textsize=tiny]{todonotes}

\begin{document}

\title{Addressing Concept Shift in Online Time Series Forecasting: Detect-then-Adapt}

\author{YiFan Zhang, Weiqi Chen, Zhaoyang Zhu, Dalin Qin, Liang Sun, Xue Wang, Qingsong Wen, Zhang Zhang, Liang Wang~\IEEEmembership{Fellow,~IEEE}, Rong Jin
\thanks{Yi-Fan Zhang, Zhang Zhang, and Liang Wang are with the State Key Laboratory of Multimodal Artificial Intelligence Systems (MAIS), Center for Research on Intelligent Perception and Computing (CRIPAC), Institute of Automation, Chinese Academy of Sciences (CASIA), Beijing 100190, China, and also with the School of Artificial Intelligence, University of Chinese Academy of Sciences (UCAS), Beijing 100049, China (e-mail: yifanzhang.cs@gmail.com).}
\thanks{Weiqi Chen, Zhaoyang Zhu, Dalin Qin, Liang Sun, and Xue Wang are with the Alibaba Group.}
\thanks{Qingsong Wen is with the Squirrel AI Group.}
\thanks{Rong Jin is with the Meta AI Group.}
}

\markboth{Journal of \LaTeX\ Class Files,~Vol.~14, No.~8, August~2021}%
{Shell \MakeLowercase{\textit{et al.}}: A Sample Article Using IEEEtran.cls for IEEE Journals}


\maketitle

\begin{abstract}
Online updating of time series forecasting models aims to tackle the challenge of concept drifting by adjusting forecasting models based on streaming data. While numerous algorithms have been developed, most of them focus on model design and updating. In practice, many of these methods struggle with continuous performance regression in the face of accumulated concept drifts over time. To address this limitation, we present a novel approach, Concept \textbf{D}rift \textbf{D}etection an\textbf{D} \textbf{A}daptation (\abbrv), that first detects drifting conception and then aggressively adapts the current model to the drifted concepts after the detection for rapid adaption. To best harness the utility of historical data for model adaptation, we propose a data augmentation strategy introducing Gaussian noise into existing training instances. It helps mitigate the data distribution gap, a critical factor contributing to train-test performance inconsistency. The significance of our data augmentation process is verified by our theoretical analysis. Our empirical studies across six datasets demonstrate the effectiveness of \abbrv in improving model adaptation capability. Notably, compared to a simple Temporal Convolutional Network (TCN) baseline, \abbrv reduces the average Mean Squared Error (MSE) by $43.9\%$. For the state-of-the-art (SOTA) model, the MSE is reduced by $33.3\%$.
\end{abstract}

\section{Introduction}
\vspace{-0.1cm}
In recent years, there has been a notable upswing in research efforts dedicated to applying deep learning techniques to time series forecasting~\cite{lim2021time,wen2022transformers}. Deep models have demonstrated exceptional proficiency, excelling not only in forecasting tasks but also in representation learning. However, predominant studies have focused on the batch learning setting, assuming the availability of the entire training dataset in advance. These approaches operate under the assumption that the relationship between input and output variables remains constant throughout the learning process. This paradigm, however, falls short in real-world scenarios where concepts are dynamic and subject to change, a phenomenon known as concept drift~\cite{tsymbal2004problem}. Concept drift implies that future data may exhibit patterns different from those observed in the past. Although re-training the model from scratch may help address the concept drift problem, it is very time-consuming. Hence, there is a growing need for the development of online training methodologies for deep forecasters, enabling incremental updates to the forecasting model with new samples. These adaptive approaches have been proved to be essential for capturing the evolving dynamics of the environment.

\begin{figure}
    \centering
    \includegraphics[width=\linewidth]{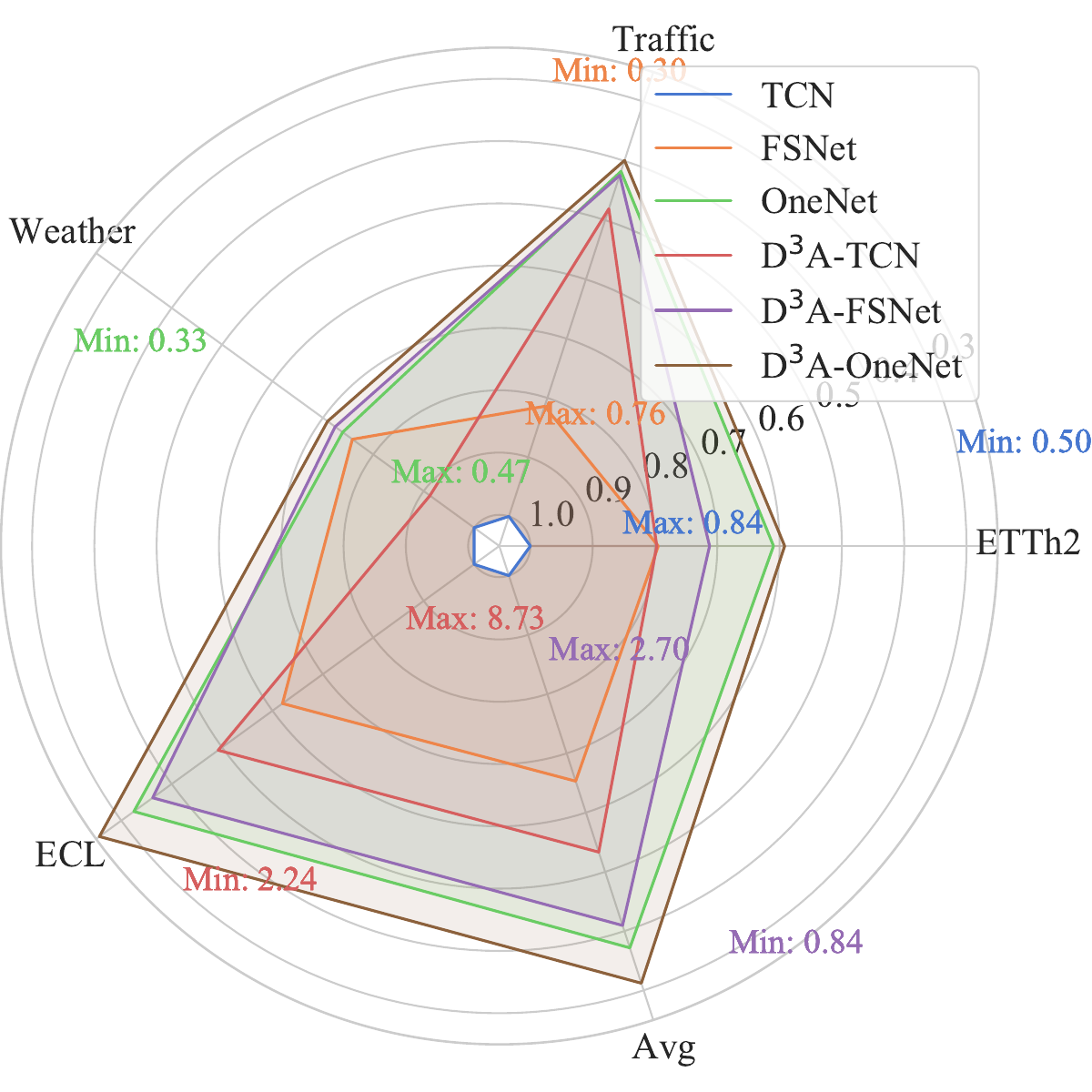}
    \vspace{-0.6cm}
    \caption{\textbf{Model performance} comparison on various tasks.}
    \label{fig:teaser}
    \vspace{-0.6cm}
\end{figure}

\begin{figure}[!t]
\centering
\subfloat[]{\includegraphics[scale=0.25]{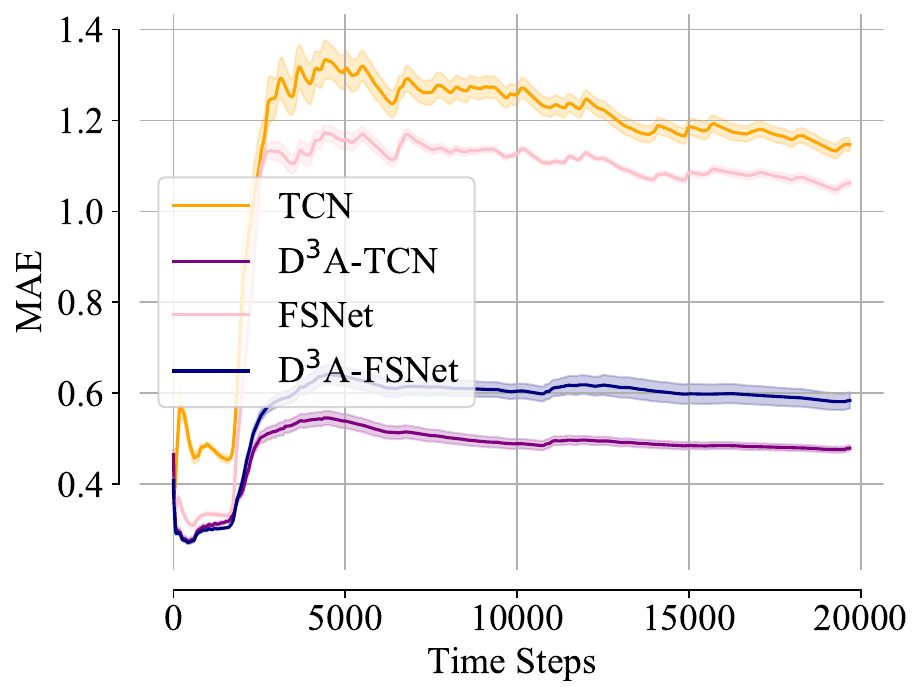}
\label{fig:ecl_mae}}
\hfil
\subfloat[]{\includegraphics[scale=0.25]{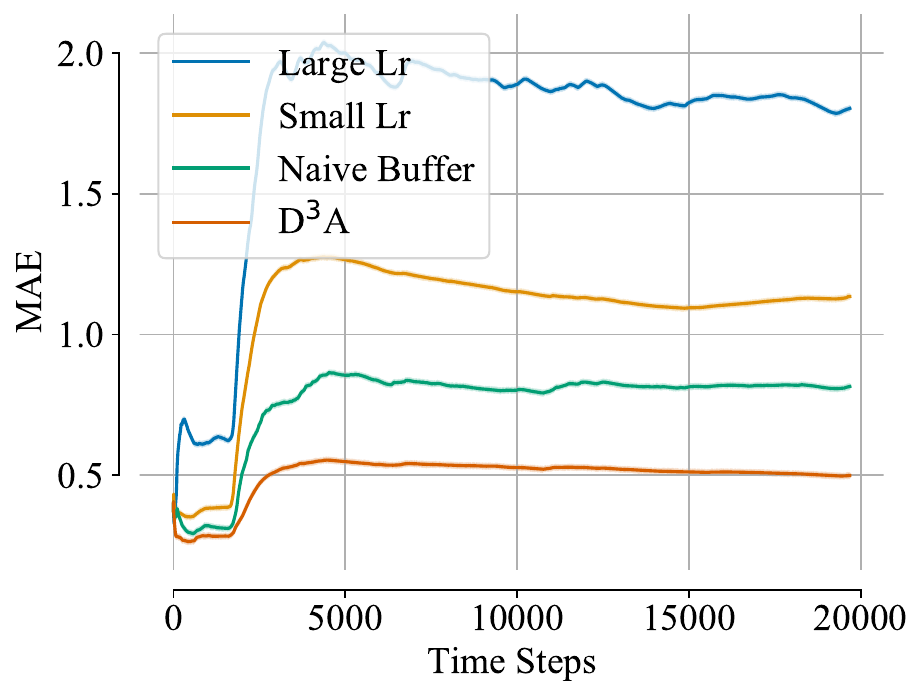}
\label{fig:ecl_mse}}
\caption{\textbf{Accumulated MAE curves} of different online learning (OL) methods on ECL dataset, the default model for (b) is FSNet.}\vspace{-0.2cm}
\label{fig:loss_compare}
\end{figure}

Engaging in online time series forecasting within real-world scenarios introduces formidable challenges marked by heightened noisy gradients, distinct from the conventional offline mini-batch training methodologies~\cite{NIPS2019_9357}. The continuous shifts in data distribution further compound the intricacies, diminishing the efficacy of models trained on historical data for contemporary predictions. Despite commendable endeavors to address these challenges through the design of sophisticated updating structures or learning objectives~\cite{pham2022learning,you2021learning}, current methodologies fall short when confronted with substantial shifts in the underlying concept. This deficiency manifests in a notable degradation of mean-average-error (MAE), as depicted in Figure~\ref{fig:loss_compare}. To delve deeper into this challenge, it is imperative to consider the limitations of individual-example-based online learning algorithms in achieving rapid adaptation. These algorithms grapple with the intricate task of balancing long-term performance and swift adaptation. A large learning rate can facilitate rapid adaptation but risks compromising long-term generalization performance by making the model overly sensitive to noise in time series data. On the contrary, a small learning rate ensures sustained long-term performance but falters in adapting quickly to distribution changes (\figurename~\ref{fig:ecl_mse}).

These observations underscore the necessity of \textbf{adapting the model aggressively upon the occurrence of a significant concept change}, as a proactive approach for adaptation becomes crucial for mitigating the inherent trade-off between rapid adjustment and sustained performance in the dynamic landscape of online time series forecasting. As a result, we introduce the framework of detect and then adapt, with one component for drift detection and one for model adaption. Unlike typical out-of-distribution detection methods that make the detection decision independently for individual data instances, our detector evaluates whether the online testing distribution significantly deviates from the historical one or surpasses the model's capacity. We make drift decision by comparing error rates. It
is grounded in the hypothesis that a statistically significant increase in the error rate serves as a trigger for a drift alarm. Our adaption module is carefully designed to address critical issues for significant concept drift. To avoid the catastrophic forgetting
problem that often arises in quick adaption, our adaption module leverages both recently acquired data and the wealth of extensive historical data for model training. 
Unlike existing replay-buffer based methods that directly mix old and new data for training and often overlook the inherent disparities between two distributions, our method makes special efforts to mitigate the performance regression caused by the distribution gap. In particular,
we incorporate data dependent Gaussian noise into the history data before they are used for model adaption, which is theoretically justified to bridge the distribution gap in the case of linear regression models and is verified in our empirical study (see Figure~\ref{fig:ecl_mse}).


Our approach, referred to as Concept \textbf{D}rift \textbf{D}etection an\textbf{d} \textbf{A}daptation (\abbrv), encompasses a novel online learning framework and training strategy. This methodology significantly alleviates the issue of sharp performance decline encountered by models in the face of substantial concept drift, as depicted in Figure~\ref{fig:loss_compare}. Our method \textbf{stands orthogonal to the majority of existing models} and when applying our updating strategy, these models all demonstrate faster adaptation to new concepts, leading to a substantial enhancement in performance (the average MSEs of different tasks are shown in~\figurename~\ref{fig:teaser}). The contributions of this paper are:

1. \textbf{Introduce a Concept Detection Framework:} Our framework monitors loss distribution drift, aiming to predict the occurrence of concept drift. This detector provides instructions for our model updating, enhancing model robustness and AI safety, particularly in high-risk tasks.

2. \textbf{Innovative Model Adaptation Strategy:} Our theoretical analysis commences with a linear case, uncovering the pivotal role played by the training-test distribution gap in the resultant model. In our pursuit of harnessing a substantial volume of historical data for enhanced model updating, we introduce a novel data augmentation strategy designed to bridge and mitigate the existing distribution gap effectively.

3. \textbf{More realistic Evaluation setting:} We observe that previous benchmarks~\cite{pham2022learning} often presume a substantial overlap in the forecasting target during testing. In this paper, we advocate for the evaluation of online time series forecasting models with delayed feedback, demonstrating a more realistic and challenging assessment.

4. \textbf{Empirical Studies:} With \textbf{six} datasets, \abbrv improves model adaptation capability across various methods. For example, compared to a simple TCN baseline, \abbrv reduces the average MSE by $43.9\%$ and MAE by $26.9\%$. For the previous SOTA model FSNet~\cite{pham2022learning}, the reductions in MSE and MAE are $33.3\%$ and $16.7\%$, respectively. In a more challenging real-world experimental setting, \abbrv consistently outperforms existing methods. For TCN, FSNet, and OneNet~\cite{zhang2023onenet}, the MSE reductions are $32\%$, $33.1\%$, and $22.2\%$, respectively.

\section{Preliminary and Related Work}\label{sec:pre}
\textbf{Concept drift.} Concepts in the real world are often dynamic and can change over time, which is especially true for scenarios like weather prediction and customer preferences. Because of unknown changes in the underlying data distribution, models learned from historical data may become inconsistent with new data, thus requiring regular updates to maintain accuracy. This phenomenon, known as concept drift~\cite{tsymbal2004problem}, adds complexity to the process of learning a model from data. In this paper, we focus on online learning for time series forecasting. Unlike most existing studies for online time series forecasting~\cite{li2022ddg,qin2022generalizing,pham2022learning} that only focus on how to online update their models, this work goes beyond parameter updating by each online instance and introduces to detect and adapt concept drift initiatively. 

\textbf{Time Series Modeling:} Time series models, pivotal in various domains, have undergone decades of development. Early data-driven approaches, such as autoregressive models like ARIMA~\cite{box1970distribution}, paved the way but grapple with challenges posed by nonlinearity and non-stationarity. Recurrent Neural Networks (RNNs) were introduced to manage sequential data, with LSTM~\cite{graves2012long} and GRU~\cite{chung2014empirical} incorporating gated structures to mitigate gradient-related issues. Attention-based RNNs~\cite{qin2017dual}, leveraging temporal attention for capturing long-range dependencies, face limitations in parallelizability and struggle with extended dependencies. Temporal Convolutional Networks~\cite{sen2019think} offer efficiency but exhibit restricted reception fields, making them less effective for handling prolonged dependencies. In recent times, transformer-based models~\cite{vaswani2017attention,wen2022transformers} have been revitalized and applied to time series forecasting. Despite extensive efforts to enhance the efficiency and potency of Transformer models~\cite{zhou2022fedformer,zhou2021informer,patchtst}, we uniquely contribute by evaluating the robustness of advanced forecasting models under concept drift. This novel approach enhances their adaptability to new distributions.

\textbf{Out-of-distribution detection and anomaly detection.} \textbf{Out-of-distribution detection methods} primarily encompass (1) classification-based methods, which derive Out-of-Distribution (OOD) scores based on classifiers trained on In-Distribution (ID) data~\cite{hendrycks2016baseline}. Examples include using the maximum softmax probability~\cite{hendrycks2016baseline} and energy~\cite{liu2020energy} as OOD scores. (2) Density-based methods model in-distribution using probabilistic models, classifying data samples in low-density regions as OOD~\cite{zong2018deep,abati2019latent,xiao2020likelihood}. (3) Distance-based methods operate under the assumption that OOD samples should be relatively distant from representations of in-distribution classes~\cite{lee2018simple,wang2022vim, zhang2023model}. \textbf{Detecting anomalies in unsupervised time series} presents a significant challenge in practical scenarios. The model must acquire informative representations of intricate temporal dynamics through unsupervised tasks and establish a distinctive criterion capable of identifying rare anomalies among numerous normal time points. Various classic methods for anomaly detection provide unsupervised paradigms, such as density-estimation methods introduced in the local outlier factor (LOF)~\cite{breunig2000lof}, clustering-based approaches exemplified by one-class SVM (OC-SVM)~\cite{scholkopf2001estimating}, and SVDD~\cite{tax2004support}. These traditional methods overlook temporal information and encounter difficulties when generalizing to unforeseen real-world scenarios. Leveraging the representation learning capabilities of neural networks, recent deep models~\cite{li2021multivariate,xu2021anomaly} have demonstrated superior performance. However, in this paper, our focus is on detecting concept shifts rather than OOD instances that differ from the ID distribution. We employ a simple sliding-window detection method to emphasize the importance of recognizing when the concept drift occurs and adapting to new environments proactively. More complex considerations involving existing OOD detection metrics can be explored in future work.

\textbf{Online time series forecasting: streaming data.} Traditional time series forecasting tasks have a collection of multivariate time series with a look-back window $L$: $(\x_i)_{i=1}^L$, where each $\x_i$ is a $M$-channel vector $\x_i=(x_i^j)_{j=1}^M$. Given a forecast horizon $H$, the target is to forecast $H$ future values $(\x_i)_{i=L+1}^{L+H}$. In real-world applications, the model builds on the historical data needs to forecast the future data, that is, given time offset $K'>L$, and $(\x_i)_{i=K'-L+1}^{K'}$, the model needs to forecast $(\x)_{i=K'+1}^{K'+H}$. Online time series forecasting~\cite{anava2013online,liu2016online,pham2022learning} is a widely used technique in real-world due to the sequential nature of the data and the frequent drift of concepts. In this approach, the learning process takes place over a sequence of rounds, where the model receives a look-back window and predicts the forecast window. The true values are then revealed to improve the model's performance in the next rounds. When we perform online adaptation, the model is retrained using the online data stream with the MSE loss over each channel: $ \mathcal{L}=\frac{1}{M}\sum_{j=1}^M \parallel \hat{x}_{K'+1:K'+H}^j - {x}_{K'+1:K'+H}^j \parallel$. 

\section{Concept Drift Detection and Adaptation}

\begin{figure}[!t] \centering \subfloat[]{\includegraphics[scale=0.25]{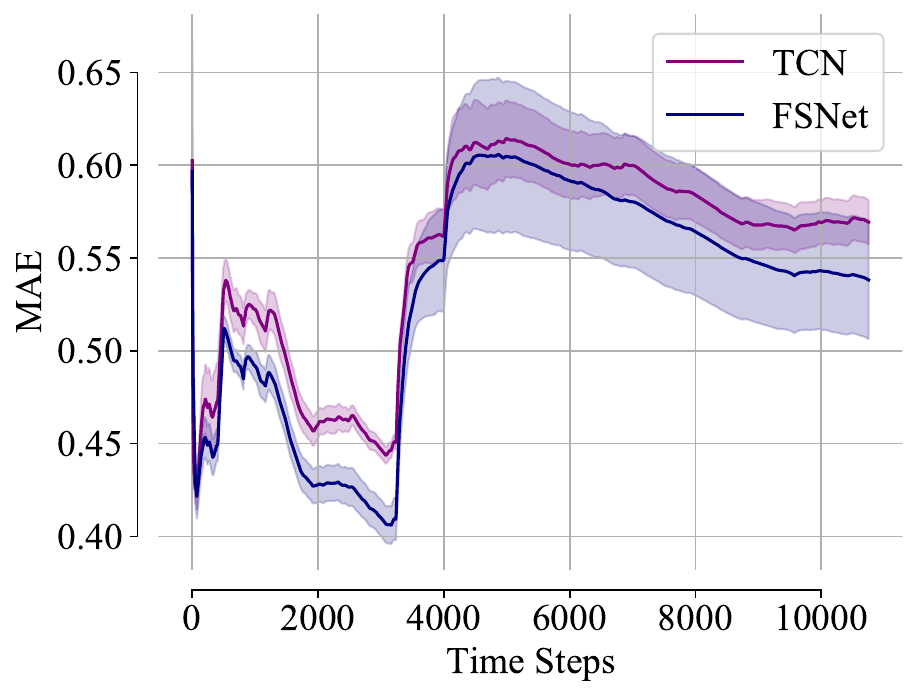}
\label{fig:etth2_naive}} \hfil \subfloat[]{\includegraphics[scale=0.25]{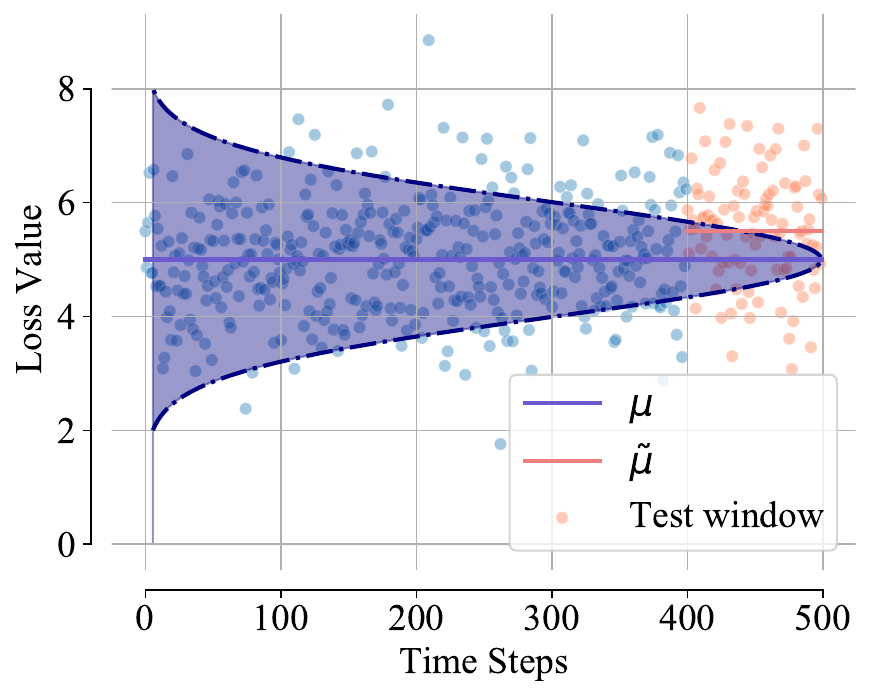}
\label{fig:detect_teaser}}\caption{(a) Accumulated MAE curves of ETTh2 and (b) Distribution of Loss Over Time, where $l_w=100$.}\vspace{-0.2cm}
\label{fig:loss_naive} \end{figure}

The current research landscape in online time series forecasting primarily revolves around model design and training strategies. However, when confronted with a substantial shift in the underlying concept, these methods grapple with mitigating the significant performance decline of the model under the new concept (see~\figurename~\ref{fig:loss_compare} and~\figurename~\ref{fig:etth2_naive}). One primary limitation with most of these online forecasting algorithms is that they gradually update their model based on one data point at each time, making it inefficient to generate enough changes in parameters when encountering significant concept drift. A recent approach, SIESTA~\cite{harun2023siesta}, suggests the need for online learning but requires a long period of time (sleep stages) for offline memory consolidation, thus not applicable to our scenario. In addition, this method requires a full fine tuning of the model, thus introducing significant overhead in computation. 

An intuitive idea to address this challenge is to perform aggressive updates of the model with new data \textbf{when a substantial concept change occurs}, allowing for rapid adaptation to the new concept and avoiding unnecessary full fine-tuning. To make this detect-then-adapt approach work in practice, we need to address three critical questions:
\begin{itemize}
\item \textit{How to determine if a significant change occurs in concept} ?, 
\item \textit{What data to use for effective model update}?, and 
\item \textit{How to reduce the computation overhead in detect-then-adapt} ?. 
\end{itemize}
In this section, we will delve into addressing these questions.

\subsection{Concept drift detection}

Unlike traditional out-of-distribution (OOD) detection or anomaly detection methods, which aim to determine whether a given data instance deviates significantly from the training or normal distribution, our focus is on assessing whether the current online testing distribution substantially differs from the training distribution or exceeds the model's capability. Our detector operates on the hypothesis that a statistically significant increase in the error rate triggers a drift alarm.

Depicted in Figure~\ref{fig:detect_teaser}, our online monitoring strategy involves tracking testing losses, assuming a normal distribution $\mathcal{N}(\mu, \sigma)$, and subsequently subjecting them to a statistical test. Specifically, employing a test window size $l_w$, we designate the latest $l_w$ losses as the test target. The mean value of losses within the test window, denoted as $\tilde{\mu}$, is computed, along with the corresponding z-score $\frac{\tilde{\mu}-\mu}{\sigma/\sqrt{|B|}}$. We then assess the confidence level at which the score of new data instances deviates from the overall distribution of test losses. Notably, in the context of a normal distribution, both significantly smaller and larger scores should trigger the drift alarm. However, in our scenario, a smaller loss signifies commendable model performance under new data instances. Consequently, the alarm is activated exclusively when $\tilde{\mu}>\mu$. In addition, we implement a detector reset every $m_t$ steps, where $m_t\geq l_w$. This design offers two distinct advantages: firstly, it prevents a scenario where outliers with high losses persist, leading to an inflated $\mu$ that does not accurately reflect the true loss distribution; secondly, this flexible approach empowers users to determine the restart time based on their preferences or the model's idleness. The corresponding pseudocode is presented in Algorithm~\ref{alg:main}.

\begin{algorithm}[tb]
    \caption{Concept Drift Detection and Adaptation}
    \label{alg:main}
\begin{algorithmic}[1]
    \STATE {\bfseries Input:} {Online multivariate time series $\mathbf{x}\in\mathbb{R}^{M\times L}$, target $\mathbf{y}\in\mathbb{R}^{M\times H}$, trained forecaster $f_\theta$}
    \STATE {\bfseries Parameters:} Window size $l_w$, a predefined confidence level $\alpha$, and a restart step $m_t\geq l_w$. \;
    \STATE {\bfseries Initialize} a memory bank $\mathcal{M}$ with FIFO updating strategy, $|\mathcal{M}|=l_w$, and loss list $B=[]$ \;
    \WHILE{$i=1,\cdots,T$}
        \STATE $\tilde{\mathbf{y}} \leftarrow f_\theta(\mathbf{x})$ \;
        \STATE $f_{\theta} \leftarrow \text{Adam}(f_\theta, \mathcal{L}(\tilde{\mathbf{y}}, \mathbf{y}))$ \textit{\color{sclgreyblue} //Update forecaster}
        \STATE $\mathcal{M} \leftarrow \mathcal{M} \cup (\mathbf{x}, \mathbf{y})$ \;
        \STATE $B \leftarrow \text{append}(B, \mathcal{L}(\tilde{\mathbf{y}}, \mathbf{y}))$ \textit{\color{sclgreyblue} //Store loss and data}
        \STATE \textit{\color{sclgreyblue} // Loss statistics.}
        \STATE $\mu \leftarrow \text{mean}(B)$, $\sigma \leftarrow \text{std}(B)$, $\tilde{\mu} \leftarrow \text{mean}(B[-l_w:])$
        \IF{($|B|>w$ \textbf{and} $\frac{\tilde{\mu}-\mu}{\sigma/\sqrt{|B|}} > \alpha$) \textbf{or} $i\%m_t==0$}
        \STATE Fully tune the model using $\mathcal{M}$;\; Reset $B=[]$
        \ENDIF
    \ENDWHILE
\end{algorithmic}
\end{algorithm}

\subsection{Better Adaptation by Reduced Distribution Gap}
Simply fully fine-tuning the model $f_\theta$ on new data points in $l_w$ may lead to limited performance. This is because (a) it could result in catastrophic forgetting of knowledge learned from historical data if we let the optimization procedure reach its optimum, and (b) the limited number of training examples in $l_w$ makes it impossible to learn an optimal model. It then prompts a crucial question: \textit{Can we leverage historical data to enhance our model adaptation?} Given historical examples stem from different distributions compared to test instances, directly using historical data for adaption could result in sub-optimal performance. We propose a data augmentation method aimed at alleviating the gap in feature distribution. It augments existing historical data with synthetically generated examples. These synthetic examples are crafted by introducing new instances-dependent Gaussian noise to historical instances. Below, we justify this approach by an analysis of a linear model, with all proofs available in appendix~\ref{proofs}.

\subsubsection{ An Analysis based on Linear Model}
Given the substantial size of our model training dataset, our analysis conveniently navigates around the intricacies introduced by finite data sampling. We employ the simplifying assumption of having direct access to the oracle of data distribution. Let $d$ denote the dimension of the input data. Our approach entails the segmentation of the overall input data into two distributions: $\P_A$ representing historical/training data and $\P_B$ representing newly observed instances. The true data distribution $\P$ emerges as a linear combination of $\P_A$ and $\P_B$, articulated as $\P = (1 - \gamma)\P_A + \gamma \P_B$, where $\gamma \in (0, 1)$ is a fixed yet unknown weight employed to linearly blend the two distributions.

Consider $y(x): \mathbb{R}^d \mapsto [0, 1]$, the deterministic output function~\footnote{For simplicity, we disregard the randomness in the output function. Nonetheless, it's relatively straightforward to extend our analysis to accommodate stochastic outputs}. Our assumption rests on the belief that $\P_A$ exhibits significantly higher quality and a more substantial volume than $\P_B$, forming the cornerstone of our analysis.
\[
    \E_{x\sim\P_A}[y(x)] \gg \E_{x\sim\P_B}[y(x)]
\]
In other words, we can almost treat $\E_{x\sim\P_B}[y(x)] \approx 0$\footnote{This does not imply that it is exactly zero, but considered negligibly small relative to $\P_A$. This assumption is a simplification aimed at emphasizing that, the historical data distribution dominates the model training.}. If we have access to the true data distribution $\P$, in case of a linear predictor, we need to solve the following optimization problem $\min\limits_{w \in \R^d} \; \L(w) = \frac{1}{2}\E_{x\sim\P}\left[|y(x) - \langle x, w\rangle|^2 \right]$. The optimal $w$ is then given by
\[
    w = \Sigma^{-1}\E_{x\sim\P}\left[y(x)x\right]
\]
\[ 
    \E_{x\sim\P}[y(x)x] = \E_{x\sim\P_A}[y(x)x] + \E_{x\sim\P_B}[y(x)x]
\]
where $\Sigma = \E_{x\sim\P}[xx^{\top}]$. Since we assume $\P_A$ has much higher influence on the model, we have 
\[ 
    \E_{x\sim\P}[y(x)x] \approx \underbrace{\E_{x\sim\P_A}[y(x)x]}_{:= z_A}
\]
Hence, we will only consider the following approximately optimal solution in comparison, i.e.
\begin{eqnarray}
    w_* = \Sigma^{-1}z_A \label{eqn:opt-solution}
\end{eqnarray}
Now, we consider the case when we only train the model with input data from distribution $\P_A$. The resulting solution is given by 
\begin{eqnarray}
    w_A = \Sigma_A^{-1}z_A \label{eqn:bias-solution}
\end{eqnarray}
where $\Sigma_A = \E_{x\sim\P_A}[xx^{\top}]$. When comparing $w_A$ to $w_*$, a discernible distinction becomes apparent in the feature correlation matrices $\Sigma_A$ and $\Sigma$. Consequently, for the linear prediction model, we observe that the discrepancy between the learned model $w_A$ and the underlying optimal model $w_*$ can be attributed to the distribution gap specifically present in the feature correlation matrices.

\subsubsection{How to Fill out the Gap?}
We will first bound how the difference between $w_*$ and $w_A$ affects the overall prediction performance in terms of the gap in feature correlations. To this end, we first define the gap matrix $\Delta(\Sigma_A)$ as
\[
\Delta(\Sigma_A) = \Sigma^{-1/2}(\Sigma - \Sigma_A)\Sigma^{-1/2} = \gamma\Sigma^{-1/2}\left(\Sigma_B - \Sigma_A\right)\Sigma^{-1/2}
\]
The following theorem allows us to bound the performance degradation of the linear predictor in terms of $|\Delta(\Sigma_A)|_2$, i.e., the spectral norm of the gap matrix.
\begin{thm}
If $|\Delta(\Sigma_A)|_2 \leq 1/2$, we have $\E_{x\sim\P}\left[\left|(w_* - w_A)^{\top}x\right|^2\right] \leq 4\L_0|\Delta(\Sigma_A)|_2^2$, where $\L_0 = z_A^{\top}\Sigma^{-1}z_A$.
\label{thm1}
\end{thm}

As indicated by Theorem~\ref{thm1}, to reduce the prediction gap with respect to the true optimal solution $w_*$, we need to come up with a different estimation of $\Sigma_A$ such that $|\Delta(\Sigma_A)|_2$ can be reduced significantly. To motivate our approach, we consider a special structure for $\Sigma_A$ and $\Sigma_B$. We assume that for the input data from distribution $\P_B$, most features are completely uncorrelated, and as a result, we approximate $\Sigma_B$ by a diagonal matrix, i.e. 
\begin{eqnarray}
    \Sigma_B = \alpha I \label{eqn:sigma-b}
\end{eqnarray}
 where $\alpha > 0$. For input distribution $\P_A$, we assume that it has $k < d$ groups of features being strongly correlated. As a result, we assume that $\Sigma_A$ can be well approximated by a combination of a diagonal matrix and a low-rank matrix, i.e. 
\begin{eqnarray}
\Sigma_A = \beta I + U\diag(\nu)U^{\top} \label{eqn:sigma-a}
\end{eqnarray}
where $\beta > 0$, $U \in \R^{d\times k}$ is an orthonormal matrix and $\nu \in \R_+^k$. The following proposition computes $|\Delta(\Sigma_A)|_2$
\begin{prop}
Assume $\alpha \geq \beta$ and $\max_{i\in [k]}\nu_i \in [\alpha - \beta, 2(\alpha - \beta)]$. We have $|\Delta(\Sigma_A)|_2 = \frac{\gamma(\alpha - \beta)}{\tau}$. 
\label{prop1}
\end{prop}

Now, we use a different $\Sigma_A$ for computing $w_A$. Define
\begin{eqnarray}
\Sigma_{A'} = \Sigma_A + \gamma I \label{eqn:sigma-a'}
\end{eqnarray}
The resulting new solution $w_{A'}$ becomes
\begin{eqnarray}
    w_{A'} = \Sigma_{A'}^{-1}z_A \label{eqn:solution-2}
\end{eqnarray}
According to Theorem 1, we have
\[
\E_{x\sim\P}\left[\left|(w_* - w_{A'})^{\top}x\right|^2\right] \leq 4\L_0|\Delta(\Sigma_{A'})|^2_2
\]
where $\Delta(\Sigma_{A'}) = \gamma\Sigma^{-1/2}(\Sigma_B - \Sigma_{A'})\Sigma^{-1/2}$.
The theorem below bounds $|\Delta(\Sigma_{A'}|_2$.
\begin{thm}
Assume $\alpha \geq \frac{2-\gamma}{3-\gamma}\beta$, and
\[
\alpha - \beta \leq |\nu|_{\infty} \leq \min\left(2(\alpha - \beta), \frac{\tau\left(3\gamma(\alpha - \beta) - \beta\right)}{\tau - \gamma(\alpha - \beta)}\right)
\]
we have $|\Delta(\Sigma_{A'})|_2 \leq |\Delta(\Sigma_A)|_2$

\label{thm2}
\end{thm}

\noindent{\bf How to implement it?} As Theorem 2 shows, if we can replace $\Sigma_A$ with $\Sigma_A' = \Sigma_A + \tau I$, we will be able to reduce the gap between historical and new data. Now, the question is how to implement it in practice. First, adding $\tau I$ to $\Sigma_A$ is equivalent to adding Gaussian noise to the input data. More specifically, for each $x$ sampled from $\P_A$, we add a random vector $u \sim \N(0, \tau I)$ to form a new sample $x' = x + u$. It is easy to verify that $\E_{x'}\left[x'(x')^{\top}\right] = \Sigma_A + \gamma I = \Sigma_{A'}$
and
$z_A' = \E_{x'}[y(x)x'] = z_A$. 
And our solution $w$ learned from the perturbed data becomes
\begin{equation}
\nonumber
\begin{aligned}
\widetilde{w} &= \left(\E_{x'}[x'(x')^{\top}]\right)^{-1}\E_{x'}[y(x)x'] \\& = \left(\Sigma_A + \gamma I \right)^{-1}z_A' = \Sigma^{-1}_{A'} z_A
\end{aligned}
\end{equation}

which is the same as the solution given in (\ref{eqn:solution-2}). 

Second, we notice that $\tau I$ is close to $\diag(\Sigma)$. Hence, we can approximate $\tau I$ by using $\diag(\Sigma)$, which requires computing the variance of individual features over distribution $\P$. Hence, in practice, we will first compute variance vector $s = \diag(\Sigma) \in \R_+^d$, with each component of $s$ corresponding to the variance of a different feature. We then construct the synthesis dataset. Let $\D = \{(x_i,y_i), i=1, \ldots, n\}$ be the original training dataset. For each $(x_i, y_i) \in \D$ from the original dataset $\D$, we sample a noise vector $u_i\sim\N(0, \diag(s))$ and create a synthesized example $(x_i' = x_i + u_i, y_i)$. Let $\D'$ be the synthesized dataset, i.e. 
\begin{equation}
  \D' = \left\{(x_i + u_i, y_i), u_i \sim\N(0, \diag(s)), i=1, \ldots, n\right\}
  \label{equ:syn}
\end{equation}

In our experimental setup, employing the original training dataset as $\mathcal{P}_A$ proves to be less intuitive due to the potential significant deviation of the concept at a later time from the training set. To address this issue, we introduce an auxiliary memory, denoted as $\mathcal{M}_{prev}$, which encompasses a substantial number of data instances from the period preceding the concept drift. Specifically, $\mathcal{M}$ may include instances $(x_i)_{i=t}^{t+l_w}$, while $\mathcal{M}_{prev}$ consists of instances $(x_i)_{t-1-l_v}^{t-1}$, where $l_v\gg l_w$. We then construct the synthesized dataset as outlined in~\ref{equ:syn}. Furthermore, we exclusively utilize data samples from $\mathcal{D}'$ as a regularization strategy to compel the model to enhance its performance on the drifted distribution. In each iteration, we randomly sample pairs $(\mathbf{x}, \mathbf{y})\in \mathcal{M}$ and $(\tilde{\mathbf{x}},\tilde{\mathbf{y}})\in \mathcal{D}'$, and subsequently compute the loss $\mathcal{L}(f_\theta(\mathbf{x}), \mathbf{y}) + \lambda \mathcal{L}(f_\theta(\tilde{\mathbf{x}}), \tilde{\mathbf{y}})$.
\begin{table*}[ht]
\caption{\textbf{MSE (top) and MAE (bottom) of various adaptation methods}. H: forecast horizon. Avg-C: the average of metric values under challenging benchmarks including ETTH2, Weather, and ECL.}\label{tab:mse}\label{tab:mae}
\extrarowheight=0.9ex
\resizebox{\linewidth}{!}{%
\centering
\begin{tabular}{lcccccccccccccccccccc}
\toprule
\rowcolor{sclgreyblue!50} & \multicolumn{3}{c}{ETTH2} & \multicolumn{3}{c}{ETTm1} & \multicolumn{3}{c}{WTH} & \multicolumn{3}{c}{Traffic} & \multicolumn{3}{c}{Weather} & \multicolumn{3}{c}{ECL} & & \\   \cline{2-4} \cline{5-7} \cline{8-10} \cline{11-13} \cline{14-16} \cline{17-19} \rowcolor{sclgreyblue!50}
\multirow{-2}{*}{MSE} & 1 & 24 & 48 & 1 & 24 & 48 & 1 & 24 & 48 & 1 & 24 & 48 & 1 & 24 & 48 & 1 & 24 & 48 &  \multirow{-2}{*}{Avg} &  \multirow{-2}{*}{Avg-C}  \\ \hline
Informer & 7.571 & 4.629 & 5.692 & 0.456 & 0.478 & 0.388 & 0.426 & 0.380 & 0.367 & - & - & - & - & - & - & - & - & - & 2.265 & 5.964 \\
OnlineTCN & 0.502 & 0.830 & 1.183 & 0.214 & 0.258 & 0.283 & 0.206 & 0.308 & 0.302 & 0.329 & 0.463 & 0.498 & 0.375 & 0.506 & 0.522 & 3.309 & 11.339 & 11.534 & 1.831 & 2.616 \\
TFCL & 0.557 & 0.846 & 1.208 & 0.087 & 0.211 & 0.236 & 0.177 & 0.301 & 0.323 & - & - & - & - & - & - & 2.732 & 12.094 & 12.110 & 2.574 & 4.925 \\
ER & 0.508 & 0.808 & 1.136 & 0.086 & 0.202 & 0.220 & 0.180 & 0.293 & 0.297 & 0.286 & 0.383 & 0.394 & 0.312 & 0.441 & 0.442 & 2.579 & 9.327 & 9.685 & 1.532 & 2.192 \\
MIR & 0.486 & 0.812 & 1.103 & 0.085 & 0.192 & 0.210 & 0.179 & 0.291 & 0.297 & - & - & - & - & - & - & 2.575 & 9.265 & 9.411 & 2.076 & 3.942 \\
DER++ & 0.508 & 0.828 & 1.157 & 0.083 & 0.196 & 0.208 & 0.174 & 0.287 & 0.294 & 0.280 & 0.374 & 0.384 & 0.308 & 0.434 & 0.435 & 2.657 & 8.996 & 9.009 & 1.478 & 2.114 \\
FSNet & 0.466 & 0.687 & 0.846 & 0.085 & 0.115 & 0.127 & 0.162 & 0.188 & 0.223 & 0.288 & 0.358 & 0.379 & 0.309 & 0.357 & 0.399 & 3.143 & 6.051 & 7.034 & 1.179 & 1.693 \\
OneNet & {0.372} & 0.559 & 0.603 & 0.087 & \textbf{0.098} & 0.133 & 0.154 & 0.181 & 0.197 & 0.223 & 0.348 & 0.381 & \textbf{0.296} & \textbf{0.314} & 0.428 & 2.481 & 2.747 & 3.279 & 0.735 & 1.059 \\ \hline \Gray
\abbrv-TCN & 0.407 & 0.753 & 0.845 & 0.091 & 0.193 & 0.201 & 0.203 & 0.329 & 0.326 & 0.304 & 0.386 & 0.405 & 0.348 & 0.456 & 0.477 & 2.347 & 5.033 & 5.507 & 1.034 & 1.439 \\ \Gray
\abbrv-FSNet & 0.381 & 0.643 & 0.767 & 0.084 & 0.111 & \textbf{0.121} & 0.160 & 0.185 & 0.219 & 0.276 & 0.336 & 0.354 & 0.308 & 0.332 & 0.377 & 2.542 & 3.162 & 3.781 & 0.786 & 1.105 \\ \Gray
\abbrv-OneNet &\textbf{ 0.362} & \textbf{0.543} & \textbf{0.584} & \textbf{0.082} & \textbf{0.098} & 0.129 & \textbf{0.150} & \textbf{0.173} & \textbf{0.191} & \textbf{0.219} & \textbf{0.337} & \textbf{0.352} & \textbf{0.296} & 0.329 & \textbf{0.371} & \textbf{2.188} & \textbf{2.122} & \textbf{2.406} & \textbf{0.608} & \textbf{0.842} \\
 \bottomrule
\end{tabular}%
}
\resizebox{\linewidth}{!}{%
\centering
\begin{tabular}{lcccccccccccccccccccc}
\toprule
\rowcolor{sclgreyblue!50} & \multicolumn{3}{c}{ETTH2} & \multicolumn{3}{c}{ETTm1} & \multicolumn{3}{c}{WTH} & \multicolumn{3}{c}{Traffic} & \multicolumn{3}{c}{Weather} & \multicolumn{3}{c}{ECL} & & \\   \cline{2-4} \cline{5-7} \cline{8-10} \cline{11-13} \cline{14-16} \cline{17-19} \rowcolor{sclgreyblue!50}
\multirow{-2}{*}{MAE} & 1 & 24 & 48 & 1 & 24 & 48 & 1 & 24 & 48 & 1 & 24 & 48 & 1 & 24 & 48 & 1 & 24 & 48 &  \multirow{-2}{*}{Avg} &  \multirow{-2}{*}{Avg-C}  \\ \hline
Informer & 0.850 & 0.668 & 0.752 & 0.512 & 0.525 & 0.460 & 0.458 & 0.417 & 0.419 & - & - & - & - & - & - & - & - & - & 0.562 & 0.757 \\
OnlineTCN & 0.436 & 0.547 & 0.589 & 0.214 & 0.381 & 0.403 & 0.276 & 0.367 & 0.362 & 0.281 & 0.361 & 0.382 & 0.235 & 0.312 & 0.326 & 0.635 & 1.196 & 1.235 & 0.467 & 0.545 \\
TFCL & 0.472 & 0.548 & 0.592 & 0.198 & 0.341 & 0.363 & 0.240 & 0.363 & 0.382 & - & - & - & - & - & - & 0.524 & 1.256 & 1.303 & 0.549 & 0.783 \\
ER & 0.376 & 0.543 & 0.571 & 0.197 & 0.333 & 0.351 & 0.244 & 0.356 & 0.363 & 0.247 & 0.299 & 0.307 & 0.16 & 0.264 & 0.281 & 0.506 & 1.057 & 1.074 & 0.418 & 0.474 \\
MIR & 0.410 & 0.541 & 0.565 & 0.197 & 0.325 & 0.342 & 0.244 & 0.355 & 0.361 & - & - & - & - & - & - & 0.504 & 1.066 & 1.079 & 0.499 & 0.694 \\
DER++ & 0.375 & 0.540 & 0.577 & 0.192 & 0.326 & 0.340 & 0.235 & 0.351 & 0.359 & 0.241 & 0.289 & 0.295 & 0.154 & 0.256 & 0.273 & 0.421 & 1.035 & 1.048 & 0.406 & 0.459 \\
FSNet & 0.368 & 0.467 & 0.515 & 0.191 & 0.249 & 0.263 & 0.216 & 0.276 & 0.301 & 0.249 & 0.284 & 0.298 & 0.161 & 0.244 & 0.292 & 0.472 & 0.997 & 1.061 & 0.384 & 0.451 \\
OneNet & 0.352 & 0.412 & 0.435 & 0.197 & \textbf{0.225} & 0.267 & 0.201 & 0.261 & 0.278 & \textbf{0.202} & 0.269 & 0.299 & 0.203 & 0.269 & 0.395 & 0.341 & 0.578 & 0.609 & 0.322 & 0.364 \\ \hline \Gray
\abbrv-TCN & 0.381 & 0.495 & 0.516 & 0.206 & 0.322 & 0.331 & 0.266 & 0.358 & 0.370 & 0.251 & 0.279 & 0.296 & 0.199 & 0.273 & 0.289 & 0.343 & 0.484 & 0.481 & 0.341 & 0.357 \\ \Gray
\abbrv-FSNet & 0.358 & 0.457 & 0.509 & {0.190} & 0.247 & \textbf{0.258} & 0.213 & 0.269 & 0.289 & 0.250 & 0.279 & 0.290 & 0.173 & 0.225 & 0.268 & 0.382 & 0.528 & 0.56 & 0.319 & 0.357 \\ \Gray
\abbrv-OneNet & \textbf{0.349} & \textbf{0.401} & \textbf{0.431} & \textbf{0.187} & \textbf{0.225} & 0.261 & \textbf{0.196} & \textbf{0.253} & \textbf{0.268} & \textbf{0.202} & \textbf{0.266} & \textbf{0.276} & \textbf{0.141} & \textbf{0.200} & \textbf{0.242} & \textbf{0.291} & \textbf{0.365} & \textbf{0.388} & \textbf{0.275} & \textbf{0.296} \\\bottomrule
\end{tabular}%
}
\end{table*}
\subsection{Importance of \abbrv}\label{sec:import}
Concept drift detection and adaptation play a crucial role in addressing real-world needs, especially in the context of on-device continual learning and inference.

\textbf{Real-World Relevance:} The paradigm of concept drift detection aligns with the practical demands of on-device continual learning and inference, as highlighted in the work by~\cite{harun2023siesta}. For instance, VR/AR headsets can leverage continual learning to define play boundaries and recognize locations in the physical world for virtual overlays. Similarly, home robots, smart appliances, and smartphones need to adapt and learn about their environments and user preferences. In these scenarios, online learning is essential, but there are also significant periods of time during which offline memory consolidation is feasible, such as when a mobile device is being charged or when its owner is asleep.

\textbf{Adaptation Amidst Device Busyness:} Even during busy device operation, the need for full-model updates becomes imperative when a substantial concept drift occurs. As depicted in Figure~\ref{fig:loss_naive}, our model's performance under a new concept might deteriorate significantly. In such cases, predictions become highly unreliable and may lead to adverse consequences, especially in high-risk tasks. Therefore, prioritizing model adjustments before deployment becomes essential, ensuring that the model is aligned with the current concept and avoiding potential negative impacts. This not only enhances the reliability of predictions but also ensures the safety and performance of intelligent systems across various applications.

\section{Experiments}\label{sec:exp}

In this section, we present compelling evidence showcasing the efficacy of \abbrv. Firstly, we demonstrate that \textbf{(1)} the proposed model achieves superior forecasting performance, reducing MSE by over $30\%$ compared to the previous SOTA model. This underscores the model's adeptness in detecting and adapting to concept drift. \textbf{(2)} we conduct comprehensive ablation studies, along with visualization and analysis, to elucidate the significance of each design choice. \textbf{(3)} we immerse ourselves in a more realistic and challenging evaluation setting, revealing \abbrv's consistent improvements over all baselines. \textbf{Lastly}, we introduce a variant of \abbrv, denoted as \abbrv$^*$, which proves more efficient than \abbrv while still outperforming the previous SOTA model by a substantial margin.

\subsection{Experimental setting}
\textbf{Datasets} We investigate a diverse set of datasets for time series forecasting. ETT~\cite{zhou2021informer}\footnote{https://github.com/zhouhaoyi/ETDataset} logs the target variable of the "oil temperature" and six features of the power load over a two-year period. We also analyze the hourly recorded observations of ETTh2 and the 15-minute intervals of ETTm1 benchmarks. Additionally, we study ECL\footnote{https://archive.ics.uci.edu/ml/datasets/ElectricityLoadDiagrams20112014} (Electricity Consuming Load), which gathers electricity consumption data from 321 clients between 2012 and 2014. The WTH\footnote{https://www.ncei.noaa.gov/data/local-climatological-data/} dataset contains hourly records of 11 climate features from almost 1,600 locations across the United States. The Weather\footnote{https://www.bgc-jena.mpg.de/wetter/} is recorded every 10 minutes for 2020 whole year, which contains 21 meteorological indicators, such as air temperature, humidity. The Traffic\footnote{http://pems.dot.ca.gov/} is a collection of hourly data from California Department of Transportation, which describes the road occupancy rates measured by different sensors on San Francisco Bay area freeways.

\textbf{Implementation Details} For all benchmarks, we set the look-back window length at 60 and vary the forecast horizon from $H = {1, 24, 48}$. We split the data into two phases: warm-up and online training, with a ratio of 25:75. We follow the optimization details outlined in~\cite{zhou2021informer} and utilize the AdamW optimizer~\cite{loshchilov2017decoupled} to minimize the mean squared error (MSE) loss. To ensure a fair comparison, we set the epoch and batch sizes to one, which is consistent with the online learning setting. We make sure that all baseline models based on the TCN backbone use the same total memory budget as FSNet, which includes three times the network sizes: one working model and two exponential moving averages (EMAs) of its gradient. For ER, MIR, and DER++, we allocate an episodic memory to store previous samples to meet this budget. For transformer backbones, we find that a large number of parameters do not benefit the generalization results and always select the hyperparameters such that the number of parameters for transformer baselines is fewer than that for FSNet. In the warm-up phase, we calculate the mean and standard deviation to normalize the online training samples and perform hyperparameter cross-validation. For different structures, we use the optimal hyperparameters that are reported in the corresponding paper.  

\textbf{Baselines of adaptation methods} We evaluate several baselines for our experiments, including methods for continual learning, time series forecasting, and online learning. Our first baseline is OnlineTCN~\cite{zinkevich2003online}, which continuously trains the model without any specific strategy. The second baseline is Experience Replay (ER)~\cite{chaudhry2019tiny}, where previous data is stored in a buffer and interleaved with newer samples during learning. Additionally, we consider three advanced variants of ER: TFCL~\cite{aljundi2019task}, which uses a task-boundary detection mechanism and a knowledge consolidation strategy; MIR~\cite{NIPS2019_9357}, which selects samples that cause the most forgetting; and DER++~\cite{buzzega2020dark}, which incorporates a knowledge distillation strategy. It is worth noting that ER and its variants are strong baselines in the online setting, as we leverage mini-batches during training to reduce noise from single samples and achieve faster and better convergence. Finally, we compare our method to FSNet~\cite{pham2022learning} and \abbr~\cite{zhang2023onenet}, which are the previous state-of-the-art online adaptation model and strategy respectively. For \abbr, we use the average version for clarity, namely the combination strategy of two branches is just averaging.

\textbf{Metrics} Because learning occurs over a sequence of rounds. At each round, the model receives a look-back window and predicts the forecast window. All models are commonly evaluated by their accumulated mean-squared errors (MSE) and mean-absolute errors (MAE); namely, the model is evaluated based on its accumulated errors over the entire learning.

\textbf{License.} All the assets (i.e., datasets and the codes for baselines) we use include an MIT license containing a copyright notice and this permission notice shall be included in all copies or substantial portions of the software.

\textbf{Environment.} We conduct all the experiments on a machine with an Intel R Xeon (R) Platinum 8163 CPU @ 2.50GHZ, 32G RAM, and four Tesla-V100 (32G) instances. All experiments are repeated $3$ times with different seeds.

\subsection{Baseline details}\label{sec:app_baseline}

We present a brief overview of the baselines employed in our experiments. 

First, OnlineTCN adopts a conventional TCN backbone~\cite{zinkevich2003online} consisting of ten hidden layers, each layer containing two stacks of residual convolution filters. 

Secondly, ER~\cite{chaudhry2019tiny} expands on the OnlineTCN baseline by adding an episodic memory that stores previous samples and interleaves them during the learning process with newer ones. 

Third, MIR~\cite{NIPS2019_9357} replaces the random sampling technique of ER with its MIR sampling approach, which selects the samples in the memory that cause the highest forgetting and applies ER to them. 

Fourthly, DER++ \cite{buzzega2020dark} enhances the standard ER method by incorporating a knowledge distillation loss on the previous logits. 

Finally, TFCL~\cite{aljundi2019task} is a task-free, online continual learning method that starts with an ER process and includes a task-free MAS-styled~\cite{aljundi2018memory} regularization. 

All the ER-based techniques utilize a reservoir sampling buffer, which is identical to that used in~\cite{pham2022learning}.

\textbf{Hyper-parameters} For the hyper-parameters of FSNet and the baselines mentioned in Section~\ref{sec:app_baseline}, we follow the setting in~\cite{pham2022learning}. Besides, we cross-validate the hyper-parameters on the ETTh2 dataset and use them for the remaining ones. In particular, we use the following configuration:

\begin{itemize}
    \item Learning rate $3e-3$ on Traffic and ECL and $1e-3$ for other datasets. Learning rate $1e-2$ for the EGD algorithm and $1e-3$ for the offline reinforcement learning block, where the selection scope is $\{1e-3, 3e-3, 1e-2, 3e-2\}$.
    \item Number of hidden layers $10$ for both cross-time and cross-variable branches, where the selection scope is $\{6, 8, 10, 12\}$. 
    \item Adapter's EMA coefficient $0.9$,  Gradient EMA for triggering the memory interaction $0.3$, where the selection scope is $\{0.1, 0.2,\dots, 1.0\}$.
    \item Memory triggering threshold $0.75$, where the selection scope is $\{0.6, 0.65, 0.7 \dots, 0.9\}$.
    \item  Episodic memory size: 5000 (for ER, MIR, and DER++), 50 (for TFCL).
\end{itemize}

\subsection{Online forecasting results}

\textbf{Cumulative Performance:} This section explores the cumulative performance of various baselines, as illustrated in Table \ref{tab:mse}, focusing on mean-squared errors (MSE) and mean-absolute errors (MAE). Notably, the \abbrv-OneNet, a novel approach that deploys OneNet with our \abbrv strategy, outperforms the majority of competing baselines across diverse forecasting horizons. Moreover, the flexibility of \abbrv enables seamless integration with advanced online forecasting methods and representation learning structures, enhancing the model's overall robustness. Interestingly, in datasets with higher forecasting error rates, \abbrv demonstrates superior performance. For instance, when applied to the challenging ECL dataset with a forecasting horizon of 48, \abbrv achieves a remarkable improvement of $52.2\%$ over TCN and a substantial improvement of $46.2\%$ over FSNet. The average MSE and MAE values of \abbrv consistently outshine those obtained using traditional online learning strategies in isolation. This highlights the pivotal role played by \abbrv in incorporating concept drift detection and bias correlation during adaptation, underscoring its significance in achieving superior predictive accuracy.

\begin{figure}[!t]
\centering
\subfloat[MSE]{\includegraphics[scale=0.25]{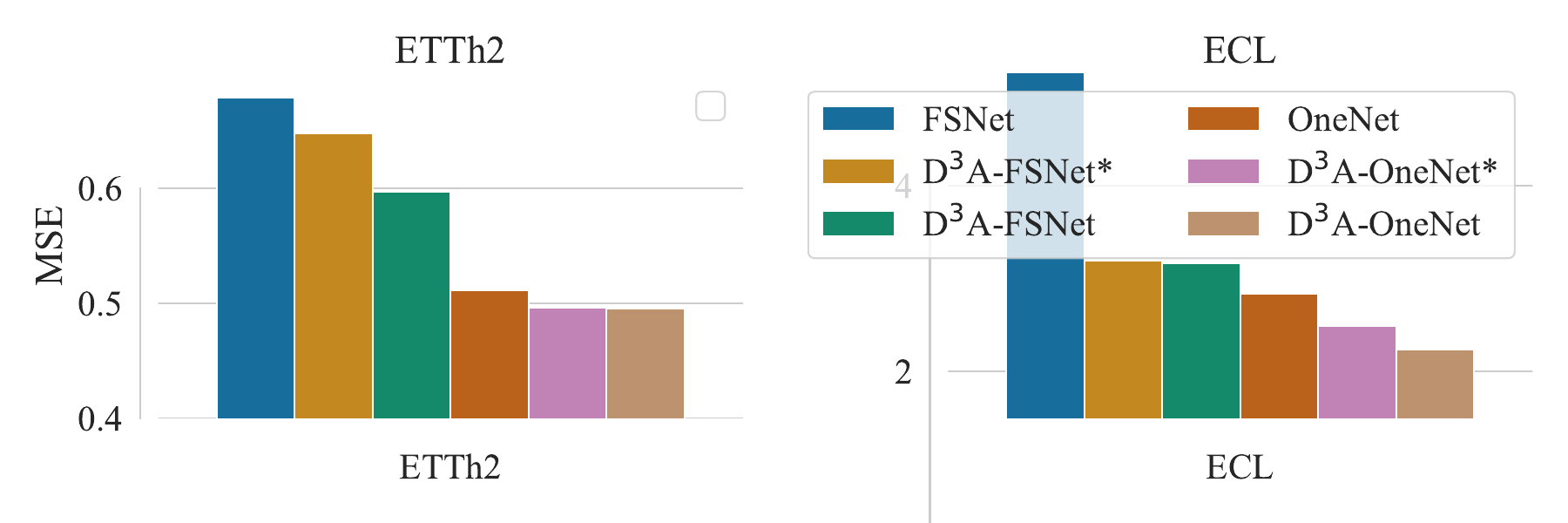}
\label{fig:minus_mse}}
\hfil
\subfloat[Inference Time]{\includegraphics[scale=0.25]{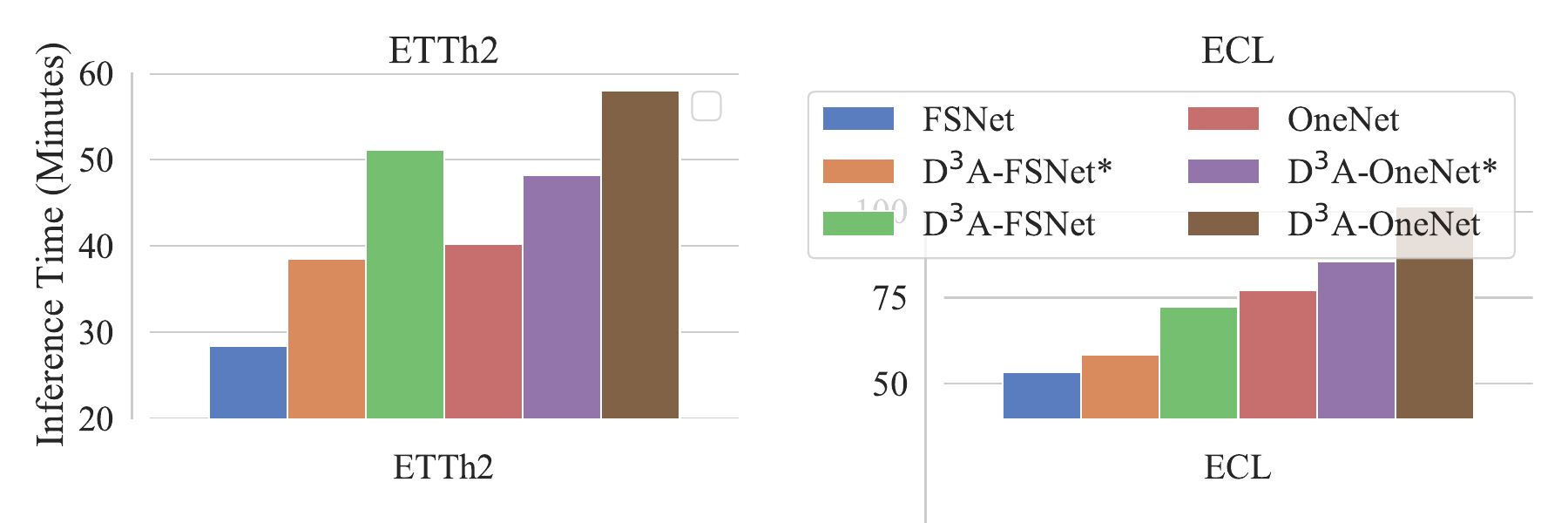}
\label{fig:minus_params}}
\caption{Comparison of MSE and Inference Time among the naive method, augmented with \abbrv, and enhanced \abbrv$^*$}
\label{fig:minus}
\end{figure}

\begin{table*}[ht]
\caption{\textbf{MSE (top) and MAE (bottom)of various adaptation methods with delayed feedback}. H: forecast horizon. Avg-C: the average of metric values under challenging benchmarks including ETTH2, Weather, and ECL.}\label{tab:mse_delay}
\extrarowheight=0.9ex
\resizebox{\textwidth}{!}{%
\begin{tabular}{lcccccccccccccccccccc}
\toprule
\rowcolor{sclgreyblue!50} & \multicolumn{3}{c}{ETTH2} & \multicolumn{3}{c}{ETTm1} & \multicolumn{3}{c}{WTH} & \multicolumn{3}{c}{Traffic} & \multicolumn{3}{c}{Weather} & \multicolumn{3}{c}{ECL} & & \\   \cline{2-4} \cline{5-7} \cline{8-10} \cline{11-13} \cline{14-16} \cline{17-19} \rowcolor{sclgreyblue!50}
\multirow{-2}{*}{MSE} & 1 & 24 & 48 & 1 & 24 & 48 & 1 & 24 & 48 & 1 & 24 & 48 & 1 & 24 & 48 & 1 & 24 & 48 &  \multirow{-2}{*}{Avg} &  \multirow{-2}{*}{Avg-C}  \\ \hline
OnlineTCN & 0.502 & 5.871 & 11.074 & 0.214 & 0.410 & 0.535 & 0.206 & 0.429 & 0.504 & 0.331 & 0.629 & 0.804 & 0.371 & 0.999 & 1.565 & 2.900 & 11.101 & 24.159 & 3.478 & 6.505 \\
ER & 0.508 & 5.461 & 17.329 & 0.086 & 0.367 & 0.498 & 0.180 & 0.373 & 0.435 & 0.286 & 0.471 & 0.567 & 0.312 & 0.896 & 1.389 & 2.388 & 8.205 & 19.528 & 3.293 & 6.224 \\
DER++ & 0.508 & 5.387 & 17.334 & \textbf{0.083} & 0.347 & 0.465 & 0.174 & 0.369 & 0.431 & 0.281 & 0.469 & 0.564 & 0.307 & 0.836 & 1.281 & 2.657 & 7.878 & 17.692 & 3.170 & 5.987 \\  \Gray
\abbrv-TCN & 0.401 & 4.752 & 9.300 & 0.091 & 0.319 & 0.381 & 0.209 & 0.383 & 0.493 & 0.308 & 0.549 & 0.800 & 0.370 & 0.826 & 1.038 & 2.341 & 6.681 & 13.582 & 2.379 & 4.366 \\
FSNet & 0.466 & 6.511 & 12.624 & 0.085 & 0.383 & 0.502 & 0.162 & 0.382 & 0.489 & 0.290 & 0.478 & 0.576 & 0.309 & 0.822 & 1.286 & 3.143 & 9.094 & 30.523 & 3.785 & 7.198 \\  \Gray
\abbrv-FSNet & 0.402 & 5.712 & 8.300 & 0.085 & 0.336 & 0.398 & 0.163 & 0.326 & 0.406 & 0.278 & 0.433 & 0.532 & 0.324 & 0.795 & 1.087 & 2.456 & 7.168 & 16.390 & 2.533 & 4.737 \\
OneNet & \textbf{0.372} & 2.807 & 5.884 & 0.087 & 0.322 & 0.368 & 0.154 & 0.323 & 0.391 & 0.223 & 0.426 & 0.514 & 0.296 & 0.722 & 1.193 & 2.472 & 5.495 & 11.363 & 1.856 & 3.400 \\  \Gray
\abbrv-OneNet & 0.377 & \textbf{2.458} & \textbf{4.323} & 0.085 & \textbf{0.285} & \textbf{0.333} & \textbf{0.152} & \textbf{0.319} & \textbf{0.390} & \textbf{0.213} & \textbf{0.391} & \textbf{0.474} & \textbf{0.286} & \textbf{0.711} & \textbf{0.964} & \textbf{2.179} & \textbf{4.865} & \textbf{7.192} & \textbf{1.444} & \textbf{2.595} \\ \bottomrule
\rowcolor{sclgreyblue!50} & \multicolumn{3}{c}{ETTH2} & \multicolumn{3}{c}{ETTm1} & \multicolumn{3}{c}{WTH} & \multicolumn{3}{c}{Traffic} & \multicolumn{3}{c}{Weather} & \multicolumn{3}{c}{ECL} & & \\   \cline{2-4} \cline{5-7} \cline{8-10} \cline{11-13} \cline{14-16} \cline{17-19} \rowcolor{sclgreyblue!50}
\multirow{-2}{*}{MAE} & 1 & 24 & 48 & 1 & 24 & 48 & 1 & 24 & 48 & 1 & 24 & 48 & 1 & 24 & 48 & 1 & 24 & 48 &  \multirow{-2}{*}{Avg} &  \multirow{-2}{*}{Avg-C}  \\ \hline
OnlineTCN & 0.436 & 1.109 & 1.348 & 0.240 & 0.511 & 0.548 & 0.276 & 0.465 & 0.508 & 0.282 & 0.414 & 0.495 & 0.229 & 0.575 & 0.787 & 0.611 & 1.141 & 1.076 & 0.614 & 0.812 \\
ER & 0.376 & 0.976 & 1.651 & 0.197 & 0.456 & 0.525 & 0.244 & 0.421 & 0.459 & 0.248 & 0.311 & 0.363 & 0.161 & 0.505 & 0.710 & 0.506 & 0.606 & 0.790 & 0.528 & 0.698 \\
DER++ & 0.375 & 0.967 & 1.644 & 0.192 & 0.443 & 0.508 & 0.235 & 0.415 & 0.456 & 0.241 & 0.309 & 0.359 & 0.154 & 0.475 & 0.671 & 0.421 & 0.591 & 0.758 & 0.512 & 0.673 \\  \Gray
\abbrv-TCN & 0.377 & 0.925 & 1.171 & 0.205 & 0.411 & 0.448 & 0.271 & 0.421 & 0.493 & 0.259 & 0.369 & 0.467 & 0.190 & 0.459 & 0.552 & 0.335 & 0.569 & 0.603 & 0.474 & 0.576 \\
FSNet & 0.368 & 0.989 & 1.472 & \textbf{0.191} & 0.468 & 0.512 & 0.216 & 0.467 & 0.518 & 0.252 & 0.320 & 0.371 & 0.160 & 0.449 & 0.665 & 0.472 & 0.923 & 1.391 & 0.493 & 0.765 \\  \Gray
\abbrv-FSNet & 0.366 & 1.000 & 1.173 & 0.192 & 0.428 & 0.461 & 0.212 & 0.385 & 0.446 & 0.254 & 0.304 & 0.356 & 0.160 & 0.429 & 0.569 & 0.371 & 0.876 & 1.112 & 0.444 & 0.673 \\
OneNet & 0.352 & 0.743 & 0.993 & 0.196 & 0.415 & 0.445 & 0.201 & 0.381 & 0.436 & 0.202 & 0.294 & 0.342 & 0.141 & \textbf{0.396} & 0.621 & 0.341 & 0.539 & 0.865 & 0.406 & 0.555 \\  \Gray
\abbrv-OneNet & \textbf{0.349} & \textbf{0.720} & \textbf{0.852} & 0.193 & \textbf{0.388} & \textbf{0.417} & \textbf{0.182} & \textbf{0.377} & \textbf{0.433} & \textbf{0.198} & \textbf{0.277} & \textbf{0.323} & \textbf{0.137} & 0.418 & \textbf{0.512} & \textbf{0.282} & \textbf{0.515} & \textbf{0.579} & \textbf{0.379} & \textbf{0.485} \\ \bottomrule
\end{tabular}%
}
\end{table*}
\textbf{Explore more efficient adaptation strategies.} As discussed in subsection~\ref{sec:import}, \abbrv plays a crucial role in addressing significant concept drift even when the model is occupied. However, we aim to enhance the efficiency of this process. Initially, we propose that, during the adaptation phase, only the linear regressor undergoes full fine-tuning until convergence, reducing backward computation time and avoiding the need for gradient storage from the resource-intensive encoder. However, a simple retraining approach proves ineffective, leading to prolonged and challenging convergence. To expedite the learning process, we employ a large learning rate ($1e-3$ by default) for warm-up, decaying it by a factor of $3$ when the average training losses in one epoch do not decrease. This strategy, termed \abbrv$^*$, achieves a favorable balance between effectiveness and efficiency. As depicted in~\figurename~\ref{fig:minus}, on two representative datasets, the inference cost of \abbrv$^*$ is significantly lower than that of \abbrv and comparable to the baseline models. Simultaneously, the performance of \abbrv$^*$ does not compromise much compared to \abbrv.

\begin{figure}[!t]
\centering
\subfloat[Channel 0.]{\includegraphics[scale=0.25]{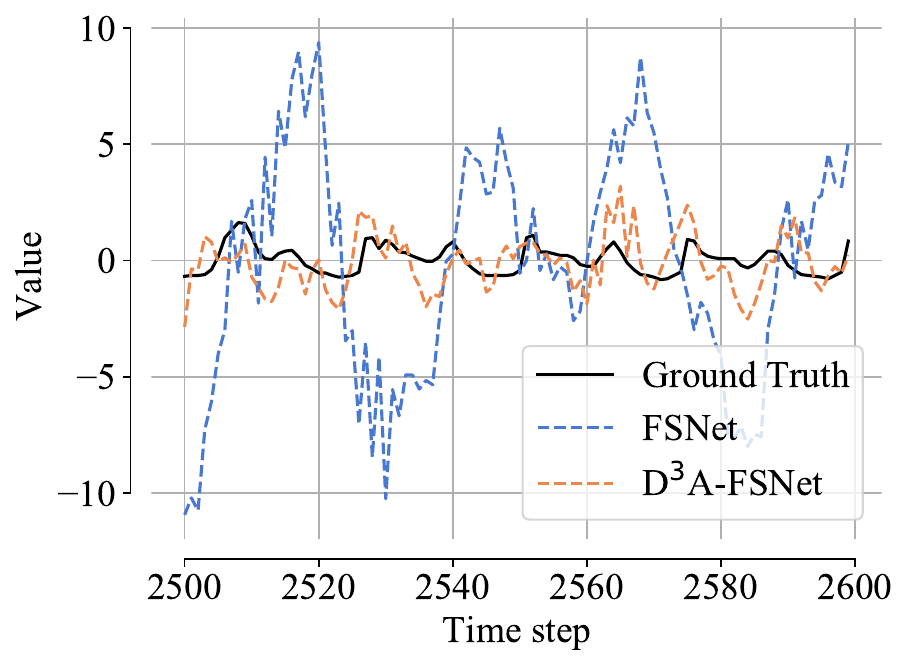}
\label{fig:visulize0}}
\hfil
\subfloat[Channel 1.]{\includegraphics[scale=0.25]{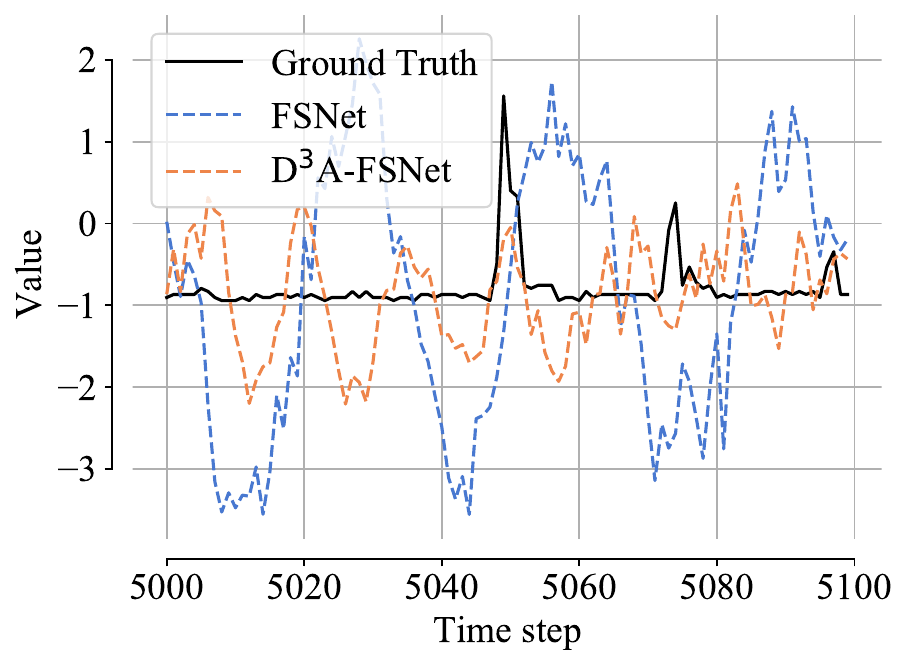}
\label{fig:visulize1}}
\caption{Visualizing the model's prediction on the ECL dataset.}\vspace{-0.2cm}
\label{fig:visualize}
\end{figure}

\textbf{Forecasting results visualization} In contrast to baseline models that encounter challenges in adapting to new concepts and yield subpar forecasting outcomes, \abbrv adeptly captures intricate patterns within time series data. As illustrated in Figure~\ref{fig:visualize}, the baseline model struggles to dynamically adjust to evolving data patterns, leading to prolonged overfitting to the existing trend. For instance, when FSNet encounters a sequence of data instances depicting an increasing trend, it tends to persistently retain this trend for an extended duration. Conversely, \abbrv actively monitors forecasting losses, prompting timely updates to the model parameters whenever they are deemed inadequate for the current data pattern. This dynamic adaptation capability enables \abbrv to navigate shifting trends more effectively, resulting in improved forecasting accuracy.

\subsection{Online forecasting results with delayed feedback}
This paper adheres to the experimental setup introduced in FSNet~\cite{pham2022learning} and \abbr~\cite{zhang2023onenet}, where the true values of each time step are disclosed to enhance the model's performance in subsequent rounds. Specifically, at time step $t$, with a look-back window $L$: $(\x_i)_{i=t}^{L+t}$ and forecasting window $H$: $(\x_i)_{i=L+1+t}^{L+H+t}$. Post-evaluation, the true values of $(\x_i)_{i=L+1+t}^{L+H+t}$ are observed and utilized for training the model. Subsequently, at time step $t+1$, the input becomes $(\x_{i})_{i=t+1}^{L+t+1}$ and the task is to forecast $(\x_i)_{i=L+2+t}^{L+H+t+1}$. Notably, there is a significant overlap in the forecasting target, with \textit{only one new point added}. It is crucial to highlight that in real-world scenarios, the true values for the forecast horizon $H$ may not be accessible until $H$ rounds later, introducing the concept of online forecasting with delayed feedback. This scenario presents increased challenges and efficiency constraints, as the model cannot undergo retraining at each round, and training is restricted to every $H$ rounds. Tables~\ref{tab:mse_delay} presents cumulative performance metrics, considering MSE and MAE. As expected, all methods exhibit reduced performance under delayed feedback compared to the traditional online forecasting setting. Notably, the SOTA method FSNet shows sensitivity to delayed feedback, particularly when $H=48$, falling behind a simple TCN baseline on specific datasets. Our proposed method, \abbrv, proves advantageous across all baselines, particularly on the challenging datasets like ETTH2, Weather, and ECL. Specifically, concerning the average MSE on the challenging subset, \abbrv reduces TCN's MSE by $\mathbf{33\%}$, while achieving performance gains of $\mathbf{34.2\%}$ and $\mathbf{23.7\%}$ for FSNet and \abbr, respectively. Additionally, \abbrv outperforms existing online learning baselines by a significant margin.

\subsection{Ablation studies and analysis}

\begin{figure*}[!t]
\centering
\subfloat[Loss weight $\lambda$.]{\includegraphics[scale=0.4]{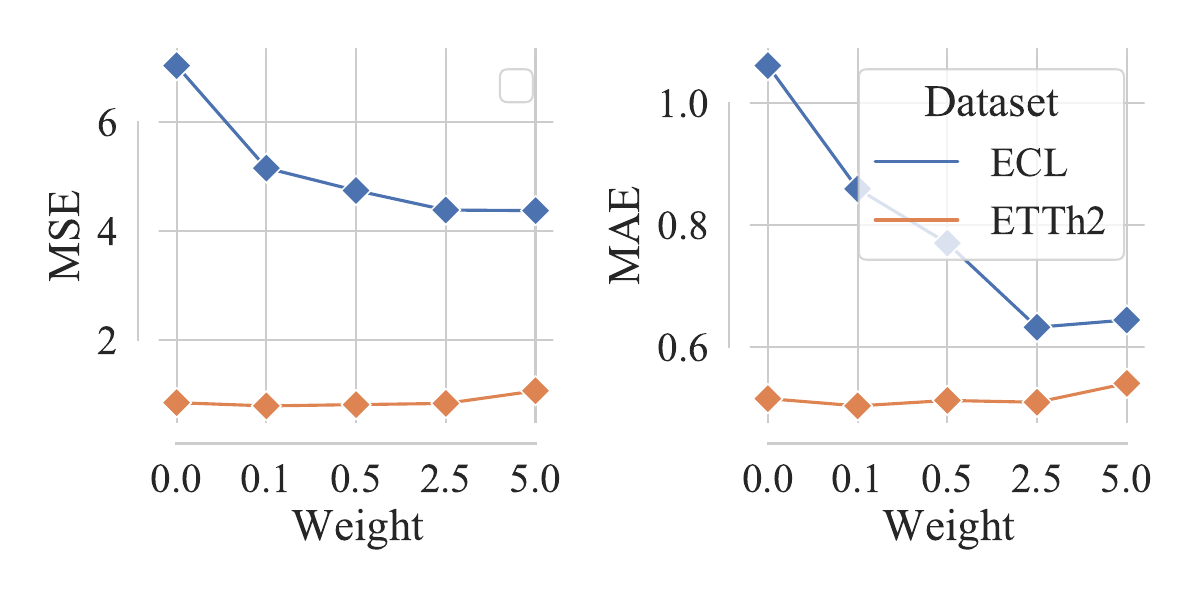}
\label{fig:ablation_weight}}
\subfloat[Memory size $l_w$.]{\includegraphics[scale=0.4]{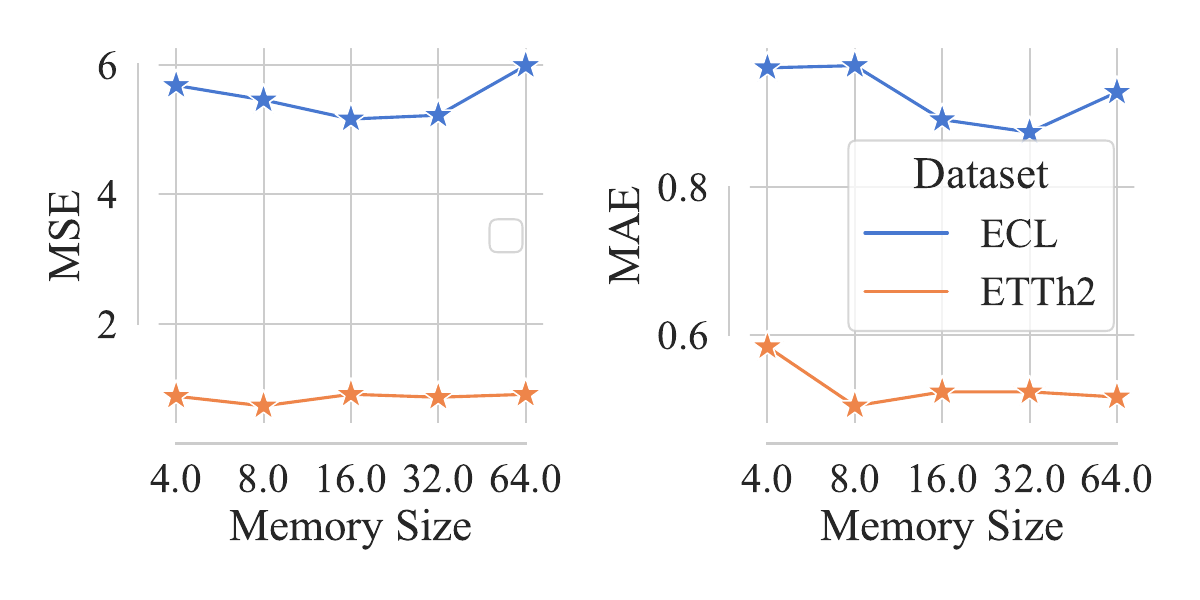}
\label{fig:ablation_memory}}
\hfil
\subfloat[Confidence level $\alpha_t$.]{\includegraphics[scale=0.4]{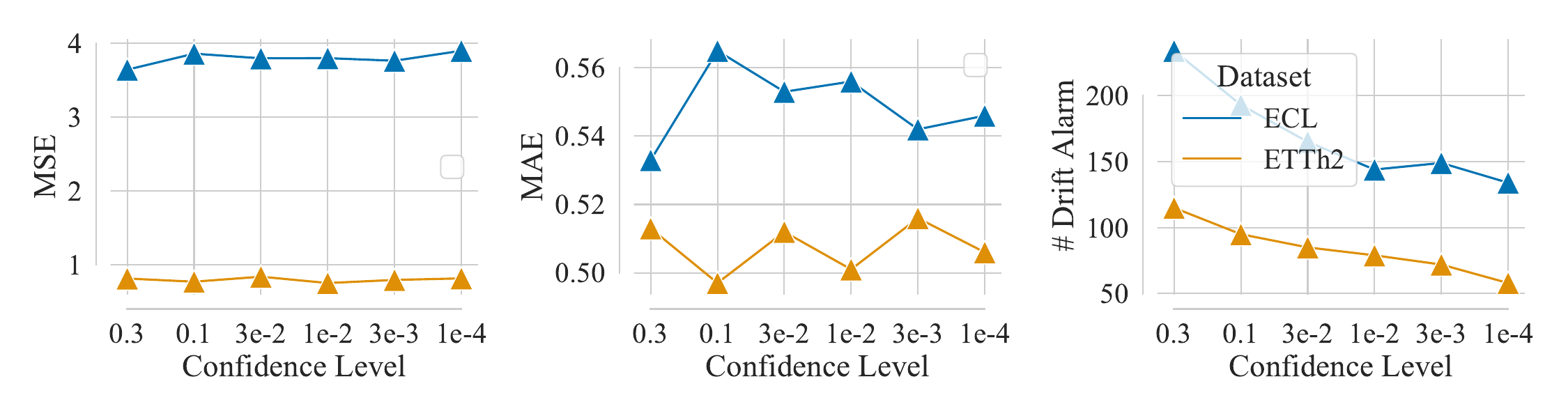}
\label{fig:ablation_confidence}}
\caption{Ablation studies of loss weight and memory size.}
\label{fig:ablation}
\end{figure*}


\textbf{Ablation studies of different hyper-parameters.}
We present ablation studies investigating the impact of different hyperparameters, specifically the loss weight $\lambda$ during adaptation and the memory size $l_w$. As depicted in Figure~\ref{fig:ablation_weight}, various datasets exhibit preferences for distinct choices. Overall, datasets with a greater number of variables tend to benefit more from the proposed augmentation strategy. This is intuitively explained by the challenging nature of learning the ECL dataset, and our augmentation strategy effectively utilizes debiased historical data to expedite the learning process. For the ETTh2 dataset, a smaller weight of $0.1$ yields optimal performance. This suggests that only a modest regularization is necessary, and training the model solely on the drifted instances allows for quick adaptation to new concepts in these datasets. Consequently, for datasets with a substantial number of variables, we default to $\lambda=2.0$, and $0.1$ otherwise. Regarding memory size, we observe a tradeoff, as a small memory size proves sensitive to concept drift and noise, while a large memory may not effectively signal when concept drift occurs. For simplicity, we choose $l_w=16$ as the default memory size for all datasets.

The confidence level is represented by the threshold $\alpha_t$ rather than the direct use of $\alpha$. Given a specific value for $\alpha_t$, we calculate the corresponding threshold $\alpha$ such that the area under the normal curve beyond this threshold aligns with the specified significance level ($\alpha_t$). Utilizing the factor $\alpha = 1 - \alpha_t / 2$ is essential for a two-tailed test. In our ablation studies, it becomes evident that as the confidence level decreases, we impose a tighter constraint on drift detection. Only when the shift in the loss distribution is significant enough does the detector trigger an alarm. Consequently, the occurrence of drift alarms and the total re-training instances are reduced. Regarding model performance, our observations indicate that more frequent retraining does not consistently enhance performance. In contrast, optimal model performance is achieved when retraining occurs judiciously based on proper data and timing considerations.

\begin{figure*}[!t]
\centering
\subfloat[Case 1. Data instance.]{\includegraphics[scale=0.3]{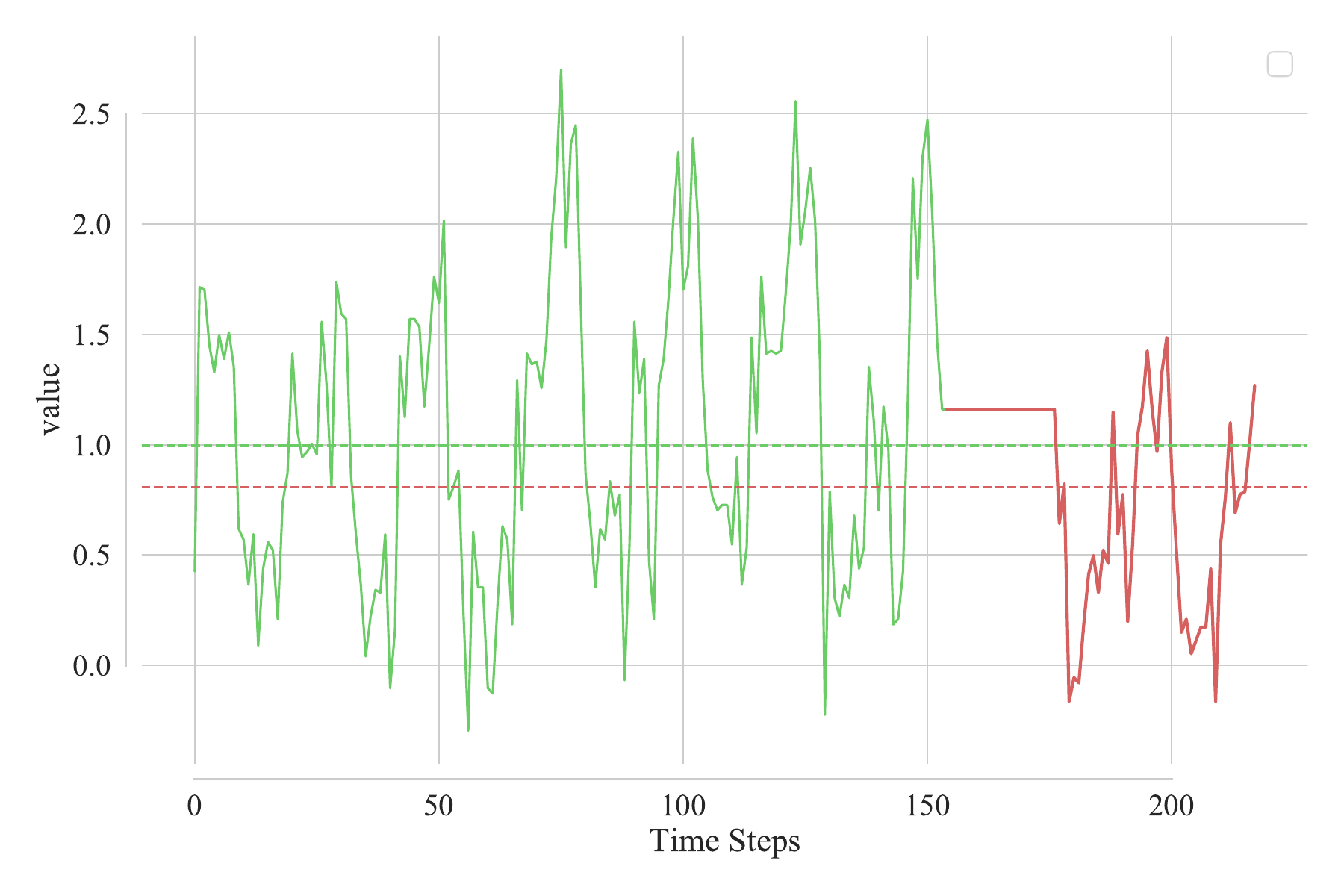}}
\subfloat[Case 1. Error.]{\includegraphics[scale=0.3]{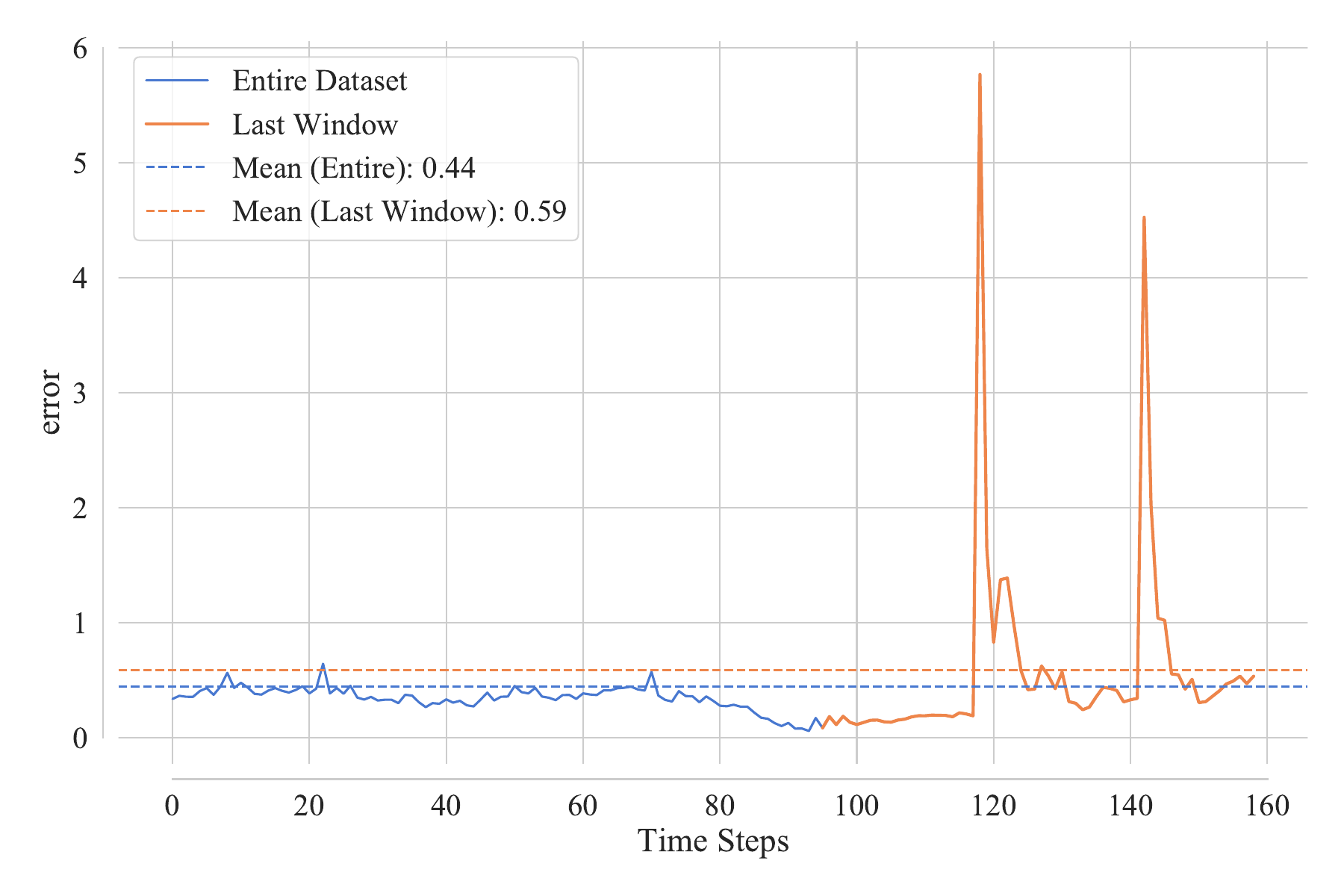}}
\hfil
\subfloat[Case 2. Data instance.]{\includegraphics[scale=0.3]{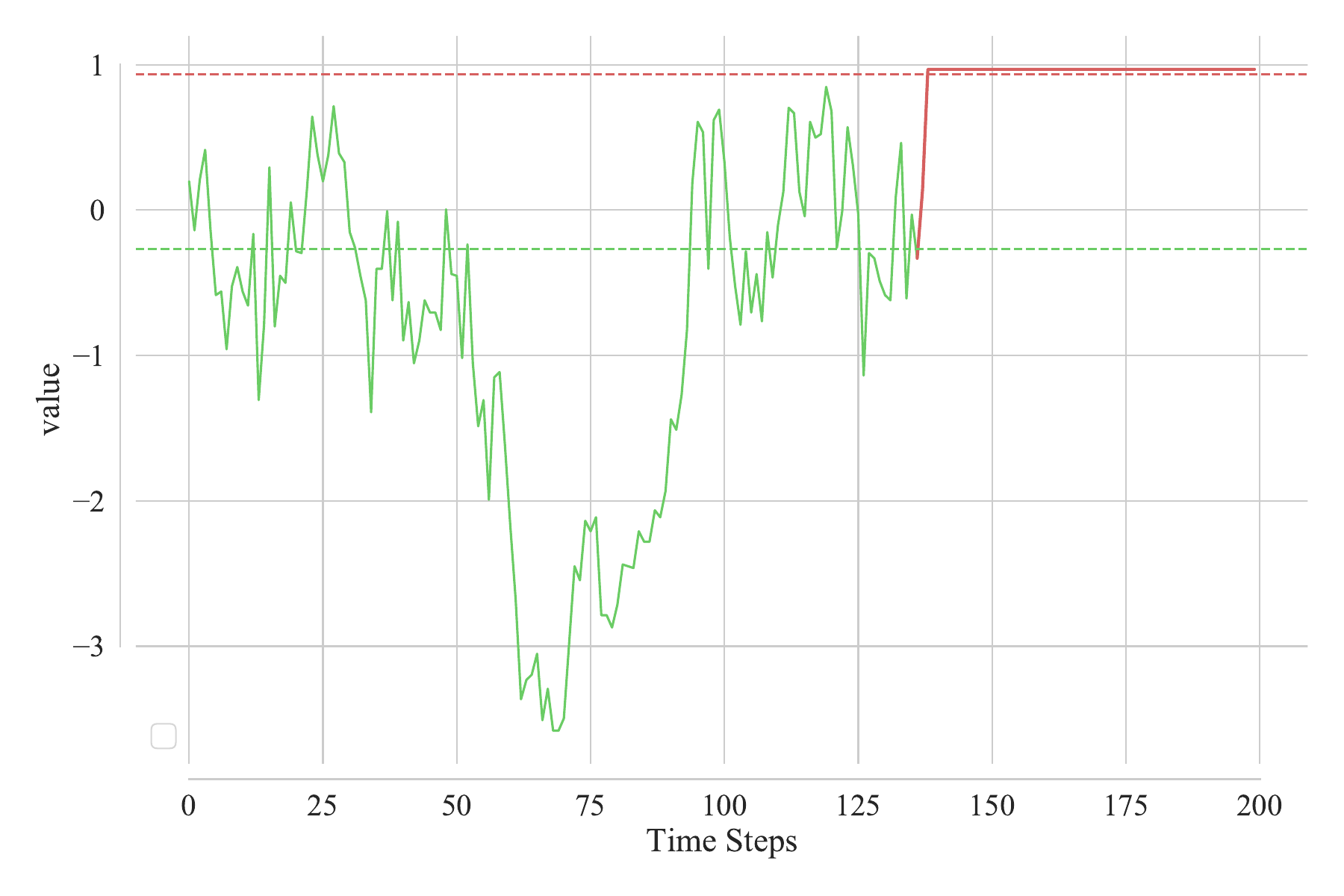}}
\subfloat[Case 2. Error.]{\includegraphics[scale=0.3]{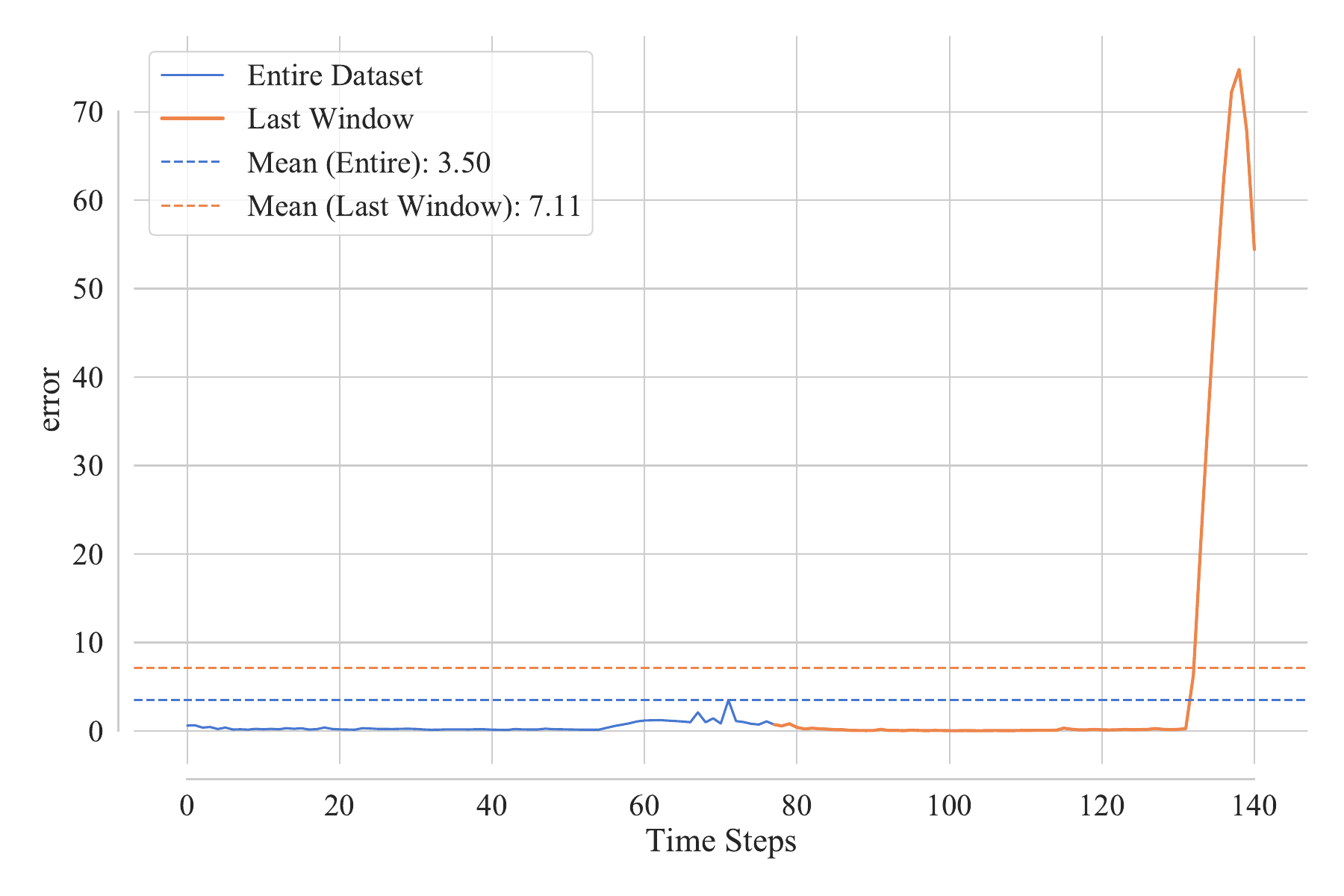}}
\hfil
\subfloat[Case 3. Data instance.]{\includegraphics[scale=0.3]{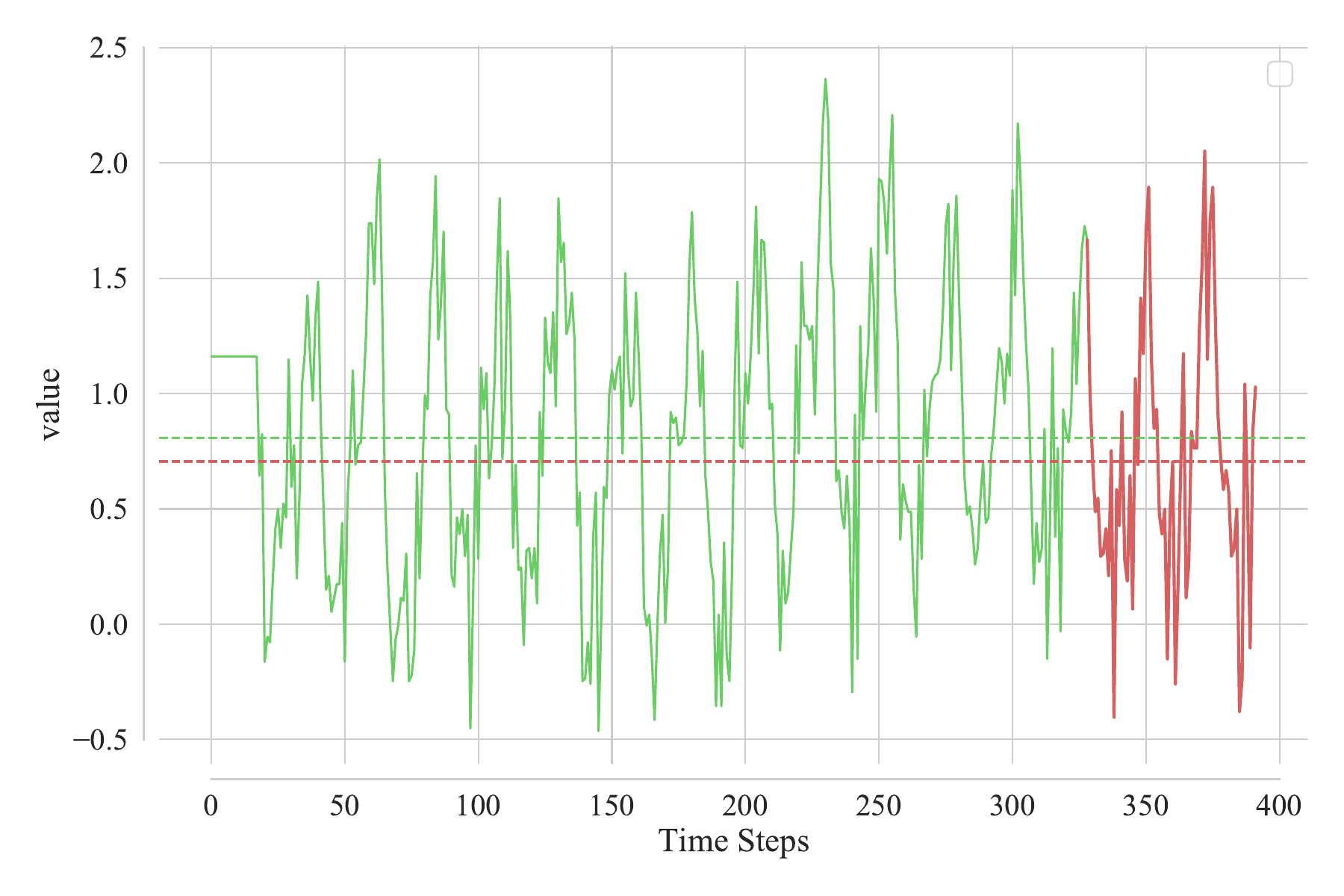}}
\subfloat[Case 3. Error.]{\includegraphics[scale=0.3]{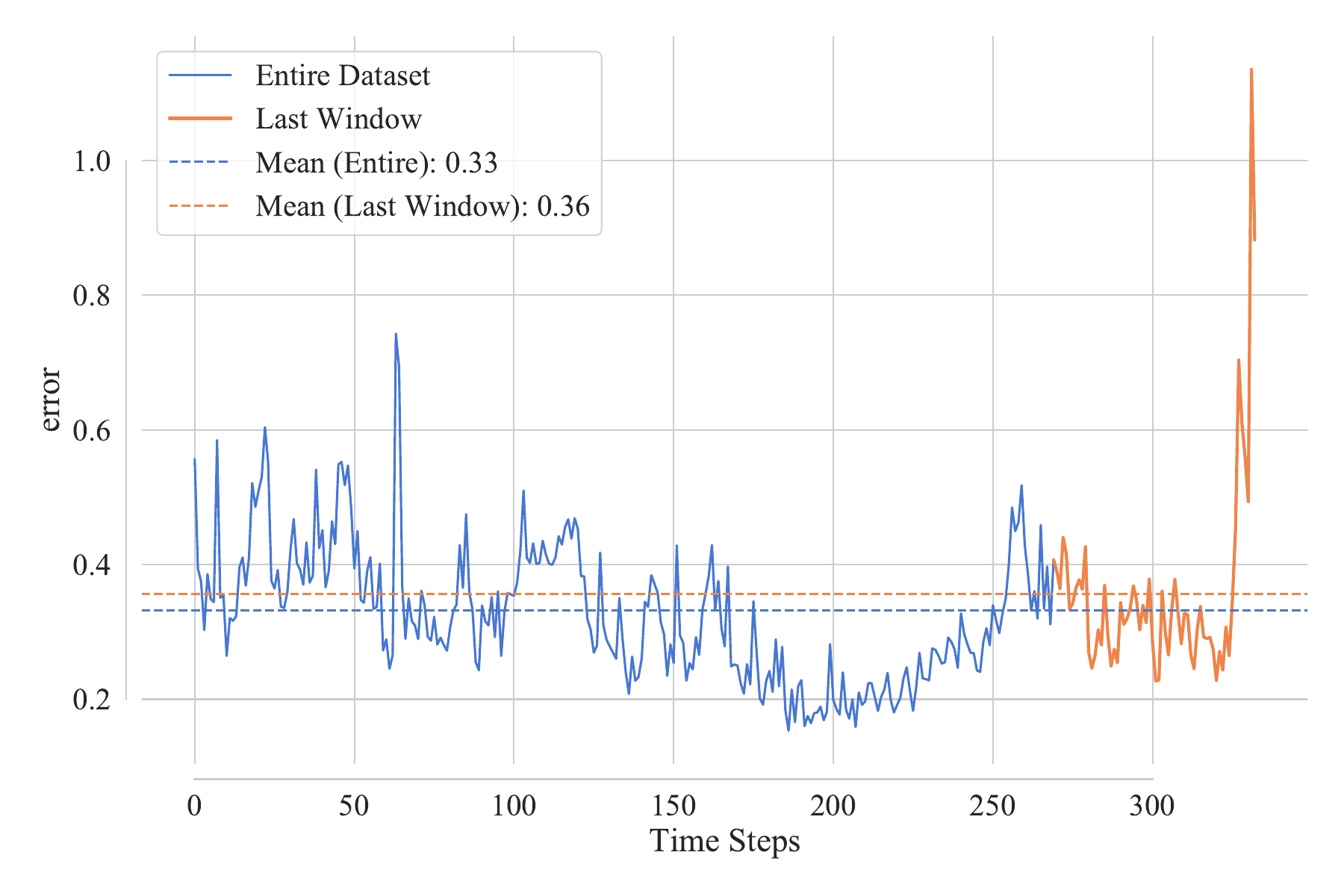}}
\caption{Visualizing instances and losses where the detector triggers a concept drift alarm in several cases, where $l_w=64$.}
\label{fig:alarm}
\end{figure*}

\textbf{Detecting Concept Drift Events: When Does the Concept Detector Trigger an Alarm?} We explore the scenarios in which our concept detector initiates a concept drift alarm by visualizing data instances and losses. The model is trained on the ECL dataset with $H$ set to $48$. As illustrated in~\figurename~\ref{fig:alarm}, in cases 1 and 2, discernible patterns emerge simultaneously in both data distribution and losses, prompting the detector to raise an alarm—an intuitive response. Intriguingly, in case 3, data patterns may not be as straightforward for human observers, yet our detector can discern a significant increase on the forecasting error, prompting the model to adapt to the new concept.

\section{Conclusion and Limitations}

This paper introduces a novel online learning framework, Concept Drift Detection and Adaptation (\abbrv), designed to tackle the challenges posed by significant concept drifts in time series forecasting. Our approach integrates a concept drift detector, which assesses the divergence between online testing and training distributions, with a data augmentation strategy aimed at enhancing adaptation using historical data. The proposed methodology showcases a substantial improvement in model adaptation capability, effectively reducing forecasting errors across a range of datasets and models. While acknowledging the introduction of additional computational burdens, we thoroughly discuss the necessity of \abbrv, particularly in high-risk tasks. Additionally, we propose a trivial method that strikes a better balance between efficiency and effectiveness.

Despite the promising outcomes, it is imperative to recognize certain limitations. The effectiveness of the proposed method may be influenced by the choice of hyperparameters and the specific characteristics of the datasets. Furthermore, the performance of the data augmentation strategy might exhibit variability in scenarios with limited historical data or under specific noise conditions. These considerations are crucial when applying \abbrv to different real-world applications. Looking ahead, exploring the adaptability of \abbrv across diverse application domains and datasets with distinct characteristics will contribute to a more comprehensive understanding of its generalizability.


\bibliography{main}

\begin{thebibliography}{10}

\bibitem{abati2019latent}
Davide Abati, Angelo Porrello, Simone Calderara, and Rita Cucchiara.
\newblock Latent space autoregression for novelty detection.
\newblock In {\em Proceedings of the IEEE/CVF conference on computer vision and pattern recognition}, pages 481--490, 2019.

\bibitem{aljundi2018memory}
Rahaf Aljundi, Francesca Babiloni, Mohamed Elhoseiny, Marcus Rohrbach, and Tinne Tuytelaars.
\newblock Memory aware synapses: Learning what (not) to forget.
\newblock In {\em Proceedings of the European conference on computer vision (ECCV)}, pages 139--154, 2018.

\bibitem{NIPS2019_9357}
Rahaf Aljundi, Eugene Belilovsky, Tinne Tuytelaars, Laurent Charlin, Massimo Caccia, Min Lin, and Lucas Page-Caccia.
\newblock Online continual learning with maximal interfered retrieval.
\newblock In {\em Advances in Neural Information Processing Systems 32}, pages 11849--11860. 2019.

\bibitem{aljundi2019task}
Rahaf Aljundi, Klaas Kelchtermans, and Tinne Tuytelaars.
\newblock Task-free continual learning.
\newblock In {\em Proceedings of the IEEE/CVF Conference on Computer Vision and Pattern Recognition}, pages 11254--11263, 2019.

\bibitem{anava2013online}
Oren Anava, Elad Hazan, Shie Mannor, and Ohad Shamir.
\newblock Online learning for time series prediction.
\newblock In {\em Conference on learning theory}, pages 172--184. PMLR, 2013.

\bibitem{box1970distribution}
George~EP Box and David~A Pierce.
\newblock Distribution of residual autocorrelations in autoregressive-integrated moving average time series models.
\newblock {\em Journal of the American statistical Association}, 65(332):1509--1526, 1970.

\bibitem{breunig2000lof}
Markus~M Breunig, Hans-Peter Kriegel, Raymond~T Ng, and J{\"o}rg Sander.
\newblock Lof: identifying density-based local outliers.
\newblock In {\em Proceedings of the 2000 ACM SIGMOD international conference on Management of data}, pages 93--104, 2000.

\bibitem{buzzega2020dark}
Pietro Buzzega, Matteo Boschini, Angelo Porrello, Davide Abati, and Simone Calderara.
\newblock Dark experience for general continual learning: a strong, simple baseline.
\newblock {\em Advances in neural information processing systems}, 33:15920--15930, 2020.

\bibitem{chaudhry2019tiny}
Arslan Chaudhry, Marcus Rohrbach, Mohamed Elhoseiny, Thalaiyasingam Ajanthan, Puneet~K Dokania, Philip~HS Torr, and Marc'Aurelio Ranzato.
\newblock On tiny episodic memories in continual learning.
\newblock {\em arXiv preprint arXiv:1902.10486}, 2019.

\bibitem{chung2014empirical}
Junyoung Chung, Caglar Gulcehre, KyungHyun Cho, and Yoshua Bengio.
\newblock Empirical evaluation of gated recurrent neural networks on sequence modeling.
\newblock {\em arXiv preprint arXiv:1412.3555}, 2014.

\bibitem{graves2012long}
Alex Graves and Alex Graves.
\newblock Long short-term memory.
\newblock {\em Supervised sequence labelling with recurrent neural networks}, pages 37--45, 2012.

\bibitem{harun2023siesta}
Md~Yousuf Harun, Jhair Gallardo, Tyler~L Hayes, Ronald Kemker, and Christopher Kanan.
\newblock Siesta: Efficient online continual learning with sleep.
\newblock {\em arXiv preprint arXiv:2303.10725}, 2023.

\bibitem{hendrycks2016baseline}
Dan Hendrycks and Kevin Gimpel.
\newblock A baseline for detecting misclassified and out-of-distribution examples in neural networks.
\newblock {\em arXiv preprint arXiv:1610.02136}, 2016.

\bibitem{lee2018simple}
Kimin Lee, Kibok Lee, Honglak Lee, and Jinwoo Shin.
\newblock A simple unified framework for detecting out-of-distribution samples and adversarial attacks.
\newblock {\em Advances in neural information processing systems}, 31, 2018.

\bibitem{li2022ddg}
Wendi Li, Xiao Yang, Weiqing Liu, Yingce Xia, and Jiang Bian.
\newblock Ddg-da: Data distribution generation for predictable concept drift adaptation.
\newblock In {\em Proceedings of the AAAI Conference on Artificial Intelligence}, volume~36, pages 4092--4100, 2022.

\bibitem{li2021multivariate}
Zhihan Li, Youjian Zhao, Jiaqi Han, Ya~Su, Rui Jiao, Xidao Wen, and Dan Pei.
\newblock Multivariate time series anomaly detection and interpretation using hierarchical inter-metric and temporal embedding.
\newblock In {\em Proceedings of the 27th ACM SIGKDD conference on knowledge discovery \& data mining}, pages 3220--3230, 2021.

\bibitem{lim2021time}
Bryan Lim and Stefan Zohren.
\newblock Time-series forecasting with deep learning: a survey.
\newblock {\em Philosophical Transactions of the Royal Society A}, 379(2194):20200209, 2021.

\bibitem{liu2016online}
Chenghao Liu, Steven~CH Hoi, Peilin Zhao, and Jianling Sun.
\newblock Online arima algorithms for time series prediction.
\newblock In {\em Proceedings of the AAAI conference on artificial intelligence}, volume~30, 2016.

\bibitem{liu2020energy}
Weitang Liu, Xiaoyun Wang, John Owens, and Yixuan Li.
\newblock Energy-based out-of-distribution detection.
\newblock {\em Advances in Neural Information Processing Systems}, 33:21464--21475, 2020.

\bibitem{loshchilov2017decoupled}
Ilya Loshchilov and Frank Hutter.
\newblock Decoupled weight decay regularization.
\newblock {\em arXiv preprint arXiv:1711.05101}, 2017.

\bibitem{patchtst}
Yuqi Nie, Nam~H. Nguyen, Phanwadee Sinthong, and Jayant Kalagnanam.
\newblock A time series is worth 64 words: Long-term forecasting with transformers.
\newblock {\em ICLR}, 2023.

\bibitem{pham2022learning}
Quang Pham, Chenghao Liu, Doyen Sahoo, and Steven~CH Hoi.
\newblock Learning fast and slow for online time series forecasting.
\newblock {\em ICLR}, 2023.

\bibitem{qin2022generalizing}
Tiexin Qin, Shiqi Wang, and Haoliang Li.
\newblock Generalizing to evolving domains with latent structure-aware sequential autoencoder.
\newblock In {\em International Conference on Machine Learning}, pages 18062--18082. PMLR, 2022.

\bibitem{qin2017dual}
Yao Qin, Dongjin Song, Haifeng Chen, Wei Cheng, Guofei Jiang, and Garrison Cottrell.
\newblock A dual-stage attention-based recurrent neural network for time series prediction.
\newblock {\em arXiv preprint arXiv:1704.02971}, 2017.

\bibitem{scholkopf2001estimating}
Bernhard Sch{\"o}lkopf, John~C Platt, John Shawe-Taylor, Alex~J Smola, and Robert~C Williamson.
\newblock Estimating the support of a high-dimensional distribution.
\newblock {\em Neural computation}, 13(7):1443--1471, 2001.

\bibitem{sen2019think}
Rajat Sen, Hsiang-Fu Yu, and Inderjit~S Dhillon.
\newblock Think globally, act locally: A deep neural network approach to high-dimensional time series forecasting.
\newblock {\em Advances in neural information processing systems}, 32, 2019.

\bibitem{tax2004support}
David~MJ Tax and Robert~PW Duin.
\newblock Support vector data description.
\newblock {\em Machine learning}, 54:45--66, 2004.

\bibitem{tsymbal2004problem}
Alexey Tsymbal.
\newblock The problem of concept drift: definitions and related work.
\newblock {\em Computer Science Department, Trinity College Dublin}, 106(2):58, 2004.

\bibitem{vaswani2017attention}
Ashish Vaswani, Noam Shazeer, Niki Parmar, Jakob Uszkoreit, Llion Jones, Aidan~N Gomez, Lukasz Kaiser, and Illia Polosukhin.
\newblock Attention is all you need.
\newblock {\em Advances in neural information processing systems}, 30, 2017.

\bibitem{wang2022vim}
Haoqi Wang, Zhizhong Li, Litong Feng, and Wayne Zhang.
\newblock Vim: Out-of-distribution with virtual-logit matching.
\newblock In {\em Proceedings of the IEEE/CVF Conference on Computer Vision and Pattern Recognition}, pages 4921--4930, 2022.

\bibitem{wen2022transformers}
Qingsong Wen, Tian Zhou, Chaoli Zhang, Weiqi Chen, Ziqing Ma, Junchi Yan, and Liang Sun.
\newblock Transformers in time series: A survey.
\newblock {\em arXiv preprint arXiv:2202.07125}, 2022.

\bibitem{xiao2020likelihood}
Zhisheng Xiao, Qing Yan, and Yali Amit.
\newblock Likelihood regret: An out-of-distribution detection score for variational auto-encoder.
\newblock {\em Advances in neural information processing systems}, 33:20685--20696, 2020.

\bibitem{xu2021anomaly}
Jiehui Xu, Haixu Wu, Jianmin Wang, and Mingsheng Long.
\newblock Anomaly transformer: Time series anomaly detection with association discrepancy.
\newblock {\em arXiv preprint arXiv:2110.02642}, 2021.

\bibitem{you2021learning}
Xiaoyu You, Mi~Zhang, Daizong Ding, Fuli Feng, and Yuanmin Huang.
\newblock Learning to learn the future: Modeling concept drifts in time series prediction.
\newblock In {\em Proceedings of the 30th ACM International Conference on Information \& Knowledge Management}, pages 2434--2443, 2021.

\bibitem{zhang2023model}
YiFan Zhang, Xue Wang, Tian Zhou, Kun Yuan, Zhang Zhang, Liang Wang, Rong Jin, and Tieniu Tan.
\newblock Model-free test time adaptation for out-of-distribution detection.
\newblock {\em arXiv preprint arXiv:2311.16420}, 2023.

\bibitem{zhang2023onenet}
YiFan Zhang, Qingsong Wen, Xue Wang, Weiqi Chen, Liang Sun, Zhang Zhang, Liang Wang, Rong Jin, and Tieniu Tan.
\newblock Onenet: Enhancing time series forecasting models under concept drift by online ensembling.
\newblock In {\em Thirty-seventh Conference on Neural Information Processing Systems (NeurIPS)}, 2023.

\bibitem{zhou2021informer}
Haoyi Zhou, Shanghang Zhang, Jieqi Peng, Shuai Zhang, Jianxin Li, Hui Xiong, and Wancai Zhang.
\newblock Informer: Beyond efficient transformer for long sequence time-series forecasting.
\newblock In {\em Proceedings of the AAAI conference on artificial intelligence}, volume~35, pages 11106--11115, 2021.

\bibitem{zhou2022fedformer}
Tian Zhou, Ziqing Ma, Qingsong Wen, Xue Wang, Liang Sun, and Rong Jin.
\newblock Fedformer: Frequency enhanced decomposed transformer for long-term series forecasting.
\newblock In {\em International Conference on Machine Learning}, pages 27268--27286. PMLR, 2022.

\bibitem{zinkevich2003online}
Martin Zinkevich.
\newblock Online convex programming and generalized infinitesimal gradient ascent.
\newblock In {\em Proceedings of the 20th international conference on machine learning (icml-03)}, pages 928--936, 2003.

\bibitem{zong2018deep}
Bo~Zong, Qi~Song, Martin~Renqiang Min, Wei Cheng, Cristian Lumezanu, Daeki Cho, and Haifeng Chen.
\newblock Deep autoencoding gaussian mixture model for unsupervised anomaly detection.
\newblock In {\em International conference on learning representations}, 2018.

\end{thebibliography}
\bibliographystyle{plain}


\clearpage
\newpage
\onecolumn
\section{Proofs of Theoretical Statements}\label{proofs}

\textit{Theorem~\ref{thm1}.} If $|\Delta(\Sigma_A)|_2 \leq 1/2$, we have
\[
\E_{x\sim\P}\left[\left|(w_* - w_A)^{\top}x\right|^2\right] \leq 4\L_0|\Delta(\Sigma_A)|_2^2
\]
where $\L_0 = z_A^{\top}\Sigma^{-1}z_A$.

\begin{proof}
We have
\begin{eqnarray*}
&   & \E_{x\sim\P}\left[\left|(w_* - w_A)^{\top}x \right|^2\right] \\
& = & \mbox{tr}\left((w_* - w_A)(w_* - w_A)^{\top} \Sigma\right) \\
& = & z_A^{\top}\left[\Sigma^{-1} - \Sigma_A^{-1}\right]\Sigma \left[\Sigma^{-1} - \Sigma_A^{-1}\right] z_A \\
& = & z_A^{\top}\Sigma^{-1/2}\left[I - \Sigma^{1/2}\Sigma_A^{-1}\Sigma^{1/2}\right]^2\Sigma^{-2}z_A \\
& = & z_A^{\top}\Sigma^{-1/2}\left[I - (I - \Delta(\Sigma_A))^{-1}\right]^2\Sigma^{-2}z_A
\end{eqnarray*}
When $|\Delta(\Sigma_A)|_2 < 1/2$, we
\[
\min \frac{u^{\top}(I + 2\Delta(\Sigma_A))(I - \Delta(\Sigma_A)) u}{|u|^2} = \min \limits_{i \in [d]} 1 + \lambda_{i}(\Delta(\Sigma_A)) - 2\lambda^2_{i}(\Delta(\Sigma_A)) \geq 1
\]
and therefore $(I + 2\Delta(\Sigma_A))(I - \Delta(\Sigma_A)) \succeq I$ or 
\[
(I - \Delta(\Sigma_A))^{-1} \preceq I + 2\Delta(\Sigma_A))
\]
Since $(I - \Delta(\Sigma_A))^{-1} \succeq I$, we have
\[
\E_{x\sim\P}\left[|(w_* - w_A)^{\top}x|^2\right] \leq 4|\Delta(\Sigma_A)|^2_2\underbrace{z_A^{\top}\Sigma^{-1}z_A}_{:=\L_0}
\]
\end{proof}

\textit{Proposition~\ref{prop1}.} Assume $\alpha \geq \beta$ and $\max_{i\in [k]}\nu_i \in [\alpha - \beta, 2(\alpha - \beta)]$. We have
\[
|\Delta(\Sigma_A)|_2 = \frac{\gamma(\alpha - \beta)}{\tau}
\]

\begin{proof}
Using the expressions for $\Sigma_B$ and $\Sigma_A$ in (\ref{eqn:sigma-b}) and (\ref{eqn:sigma-a}), we can rewrite $\Delta$ as
\begin{eqnarray*}
\Delta(\Sigma_A) & = & \gamma\left(\tau I + \gamma U\diag(\nu)U^{\top}\right)^{-1/2}\left((\alpha - \beta)I - U\diag(\nu)U^{\top}\right)\left(\tau I + \gamma U\diag(\alpha)U^{\top}\right)^{-1/2} \\
& = & - I + \gamma\left(\alpha - \beta + \frac{\tau}{\gamma}\right)\left(\tau I + \gamma U\diag(\nu)U^{\top}\right)^{-1}
\end{eqnarray*}
where $\tau = (1-\gamma)\beta + \gamma\alpha$. 
Since $\alpha > \beta$, we have
\[
|\Delta(\Sigma_A)|_2 = \max\left(\frac{\gamma(\alpha - \beta)}{\tau}, 1 - \frac{\gamma(\alpha - \beta) + \tau}{\tau + \gamma |\nu|_{\infty}}\right)
\]
where $|\nu|_{\infty}$ is infinity norm of vector $\nu$. When $|\nu|_{\infty} \in [\alpha - \beta, 2(\alpha - \beta)]$, it is easy to verify that
\[
1 - \frac{\gamma(\alpha - \beta) + \tau}{\tau + \gamma|\nu|_{\infty}} = \frac{\gamma\left(|\nu|_{\infty} - [\alpha - \beta]\right)}{\tau + \gamma |\nu|_{\infty}} \leq \frac{\gamma(\alpha - \beta)}{\tau + \gamma|\nu|_{\infty}} < \frac{\gamma(\alpha - \beta)}{\tau}
\]
and therefore
\[
    |\Delta(\Sigma_A)|_2 = \frac{\gamma(\alpha - \beta)}{\tau}
\]
\end{proof}

\textit{Theorem~\ref{thm2}.}
Assume 
\[
\alpha \geq \frac{2-\gamma}{3-\gamma}\beta, \quad \alpha - \beta \leq |\nu|_{\infty} \leq \min\left(2(\alpha - \beta), \frac{\tau\left(3\gamma(\alpha - \beta) - \beta\right)}{\tau - \gamma(\alpha - \beta)}\right)
\]
we have
\[
|\Delta(\Sigma_{A'})|_2 \leq |\Delta(\Sigma_A)|_2
\]

\begin{proof}
Plugging the expression for $\Sigma_{A'}$ (i.e.~\ref{eqn:sigma-a'}) into the expression for $\Delta(\Sigma_{A'}$, we have
\begin{eqnarray*}
\Delta(\Sigma_{A'}) & = & \left(\tau I + \gamma U\diag(\nu)U^{\top}\right)^{-1/2}\left(\gamma (\alpha - \beta) I - \left[\tau I + U\diag(\nu)U^{\top}\right]\right)\left(\tau I + \gamma U\diag(\alpha)U^{\top}\right)^{-1/2} \\
& = & \left(\gamma(\alpha - \beta) + (1-\gamma)\tau\right)\left(\tau I + \gamma U\diag(\nu)U^{\top}\right)^{-1} - 1
\end{eqnarray*}
Since $|\nu|_{\infty} > \alpha - \beta$, we have
\[
|\Delta(\Sigma_{A'})|_2 = 1 - \frac{\gamma(\alpha - \beta) + (1 - \gamma)\tau}{\tau + \gamma |\nu|_{\infty}}
\]
Since
\[
|\nu|_{\infty} \leq \frac{\tau(2[\alpha -\beta] - \tau)}{\tau - \gamma(\alpha - \beta)}
\]
we have
\[
|\Delta(\Sigma_{A'})|_2 < \frac{\gamma(\alpha - \beta)}{\tau} = |\Delta(\Sigma_A)|_2
\]
leading to less discrepancy in prediction between $w_{A'}$ and $w_*$.
\end{proof}
 




\vfill

\end{document}